\pdfoutput=1

\documentclass[11pt]{article}

\usepackage[final]{coling}

\usepackage{times}
\usepackage{latexsym}

\usepackage[T1]{fontenc}

\usepackage[utf8]{inputenc}

\usepackage{microtype}

\usepackage{inconsolata}

\usepackage{graphicx}

\usepackage{color, colortbl} 
	
\definecolor{Gray}{gray}{0.9}
\usepackage{svg}
\usepackage{subcaption}
\usepackage{tabularx}
\usepackage{multirow}
\usepackage{geometry}
\usepackage{amssymb}
\usepackage{hyperref}
\usepackage{wrapfig}
\usepackage{fontawesome}
\usepackage{booktabs}
\usepackage{algorithm}
\usepackage{algpseudocode}
\usepackage{amsmath} 
\usepackage{comment}

\title{Analysing Zero-Shot Readability-Controlled Sentence Simplification}

\author{
\textbf{Abdullah Barayan\textsuperscript{1,2}},
\textbf{Jose Camacho-Collados\textsuperscript{1}} \and
\textbf{Fernando Alva-Manchego\textsuperscript{1}}
\\
\textsuperscript{1} School of Computer Science and Informatics, Cardiff University, UK\\
\textsuperscript{2} Faculty of Computing and Information Technology, King Abdulaziz University,\\ Jeddah, Saudi Arabia
\\
\texttt{\{barayanas,camachocolladosj,alvamanchegof\}@cardiff.ac.uk} }

\begin{document}
\maketitle
\begin{abstract}
    Readability-controlled text simplification (RCTS) rewrites texts to lower readability levels while preserving their meaning. 
RCTS models often depend on parallel corpora with readability annotations on both source and target sides.
Such datasets are scarce and difficult to curate, especially at the sentence level. 
To reduce reliance on parallel data, we explore using instruction-tuned large language models for zero-shot RCTS.
Through automatic and manual evaluations, we examine: (1) how different types of contextual information affect a model's ability to generate sentences with the desired readability, and (2) the trade-off between achieving target readability and preserving meaning. 
Results show that all tested models struggle to simplify sentences (especially to the lowest levels) due to models' limitations and characteristics of the source sentences that impede adequate rewriting. 
Our experiments also highlight the need for better automatic evaluation metrics tailored to RCTS, as standard ones often misinterpret common simplification operations, and inaccurately assess readability and meaning preservation.

\end{abstract}

\section{Introduction}

\begin{figure}[tb]
    \centering
        \includegraphics[width=\linewidth]{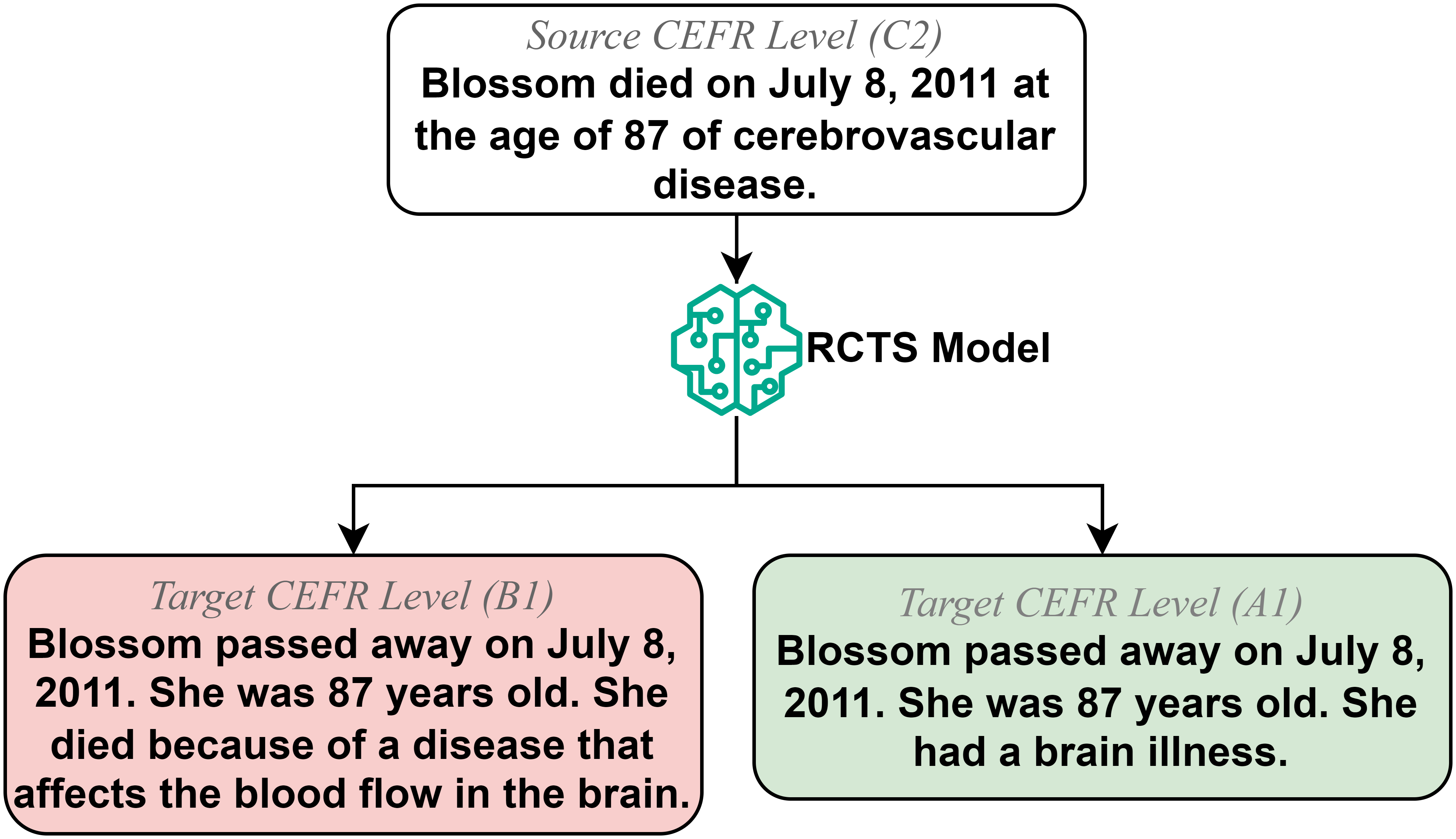} 

\caption{Example of an RCTS model rewriting a text with CEFR level C2 to either A1 or B1 levels.}
\label{fig:RCTSexamples}
\end{figure}

Text simplification (TS) consists of rewriting a text into an easier-to-understand version while preserving its meaning \citep{alva-manchego-etal-2020-data,TSSurvey}. 
Readability-controlled TS (RCTS), in particular, aims to generate simplifications in specified lower readability levels \citep{scarton-specia-2018-learning}.
This is useful when creating language learning materials for children~\citep{gala-etal-2020-alector,Javourey-Drevet_Dufau_François_Gala_Ginestié_Ziegler_2022} and non-native speakers \citep{gradedarticle} to assist with reading comprehension.
Figure~\ref{fig:RCTSexamples} shows how an RCTS model can rewrite a text from a high readability level to different lower ones.

Most methods for sentence-level RCTS rely on fully supervised training approaches \citep{nishihara-etal-2019-controllable,spring-etal-2021-exploring,rios-etal-2021-new,yanamoto-etal-2022-controllable,zetsu-etal-2022-lexically,agrawal-carpuat-2023-controlling}, and are hindered by the scarcity of high-quality annotated datasets that include original-simplified sentence pairs alongside readability levels.
Parallel datasets such as Newsela~\citep{xu-etal-2015-problems} and OneStopEnglish~\citep{vajjala-lucic-2018-onestopenglish} contain readability annotations only at the document level, which cannot be directly transferred to individual paragraphs or sentences in each text.
To address this, we investigate using instruction-tuned large language models (LLMs) for RCTS on English sentences without parallel annotated data (i.e.\ in a zero-shot setting).
In particular, we use the CEFR-SP dataset ~\citep{arase-etal-2022-cefr}, since the CEFR standard offers multiple readability targets, which challenge controlled generation models. 
We exploit LLMs' ability to follow natural language instructions by providing different types of contextual information about the desired readability, including descriptions of readers' skills at the target CEFR level and examples of sentences with the required readability.

We add to existing work on LLMs for RCTS \citep{imperial-tayyar-madabushi-2023-flesch,farajidizaji2023possible,agrawal-carpuat-2023-controlling} by: (1) analysing models' performance at the sentence level rather than the passage level; (2) examining a higher number of LLMs (with a total of 8 models), and (3) measuring not just a model's ability to meet the target readability, but also how this affects preserving the original meaning using both automatic metrics and human assessment.
We find that:

\begin{itemize}

    \item Generating simplifications at the specified readability levels was difficult for all models, especially when the gap between the source and target readability levels was considerable;

    \item Providing specific information about the target readability does not have a consistent impact across all models;

    \item Improving evaluation metrics for RCTS is crucial, as current methods often overlook key simplification operations that are well regarded by human judges. Consequently, automatic readability predictors tend to overestimate text complexity, and semantic similarity metrics penalize effective paraphrasing.
\end{itemize}

\section{Sentence-Level RCTS Approach}
\label{sec:approach}

We leverage the instruction-tuned LLMs' ability to follow natural language instructions, which we use to specify the desired readability levels and provide contextual details about them.

\subsection{Target Readability Levels}
\label{sec:target-levels}
As our target readers, we chose learners of English as a second language.
One of the most recognised metrics for measuring language learning is the Common European Framework of Reference for Languages (CEFR),\footnote{\url{https://www.coe.int/en/web/common-european-framework-reference-languages}} which categorizes language proficiency on a six-point reference scale: A1, A2, B1, B2, C1 and C2, denoting increasing levels of difficulty.
For our study, sentences in CEFR levels C2, C1 and B2 are considered as ``difficult'' for our readers, and we analyse the LLMs' ability to rewrite them to ``simpler'' levels B1, A2, and A1.

\subsection{Contextual Information}
\label{sec:Prompts}

We explore how contextual details related to target readability levell affect LLMs' ability to control readability in generated simplifications, particularly examining two types of information:
\begin{itemize}
    \item \textbf{CEFR Descriptor:} specifies what readers at the target CEFR level can understand, according to the overall reading comprehension scale in the communicative language activities CEFR Descriptor Scheme.\footnote{Each CEFR level descriptor can be found in Appendix \ref{app:CEFRLevelsDescriptors}.}
    \item \textbf{CEFR Examples:} examples of sentences that readers at the desired CEFR level can understand. For each level, we choose three random sentences from the training split.\footnote{Studies on the effect of the number of examples or smarter strategies for selecting them are left for future work.}
\end{itemize}
We designed four prompts based on the same general instruction template that details the task and target CEFR level, but with variations in the additional information provided: (P1) target, (P2) target+descriptor, (P3) target+examples, and (P4) target+descriptor+examples. 
Following previous work in TS \citep{kew-etal-2023-bless}, we adopt the detailed prompting style that describes the task.
The specific prompts used can be found in Appendix~\ref{app:appendix-prompts}.

\subsection{Instruction-Tuned LLMs} 

We investigate 8 instruction-tuned LLMs with diverse sizes and two types of architecture: an encoder-decoder and a standalone decoder, including open-weight and closed-weight models. For the open-weight models, we use LLMs in the 80M-8B range: \textbf{Flan-T5} \citep{chung2022scaling} in four size variants Small (80M), Base (250M), Large (780M), and XL (3B); \textbf{Llama-3-instruct-8B} \citep{llama3modelcard}; \textbf{OpenChat\_3.5-7B} \citep{wang2023openchat}; and Mistral-7B-Instruct \citep{jiang2023mistral}. 
For closed models, we use \textbf{GPT-3.5-Turbo} and \textbf{GPT-4-Turbo} from OpenAI.\footnote{\url{https://platform.openai.com/docs/models}} More information on the models can be found in Appendix~\ref{app:models-descriptions}.

\section{Evaluation Setup}
\label{sec:experimental-setup}

\subsection{Test Set}

For our study, we relied on the CEFR-based Sentence Profile corpus~\citep[CEFR-SP,][]{arase-etal-2022-cefr}. 
This is a sentence-level readability assessment dataset, containing 17k English sentences sourced from Newsela-Auto and Wiki-Auto (as collected by \citet{jiang-etal-2020-neural}), and SCoRE~\citep{chujo2015corpus}.
As such, its sentences belong to three different domains: news, encyclopedic (Wikipedia) and English language learning, respectively.
Each sentence in the dataset was annotated by two English-education experts with a corresponding CEFR standard scale.
 
To identify ``complex'' instances for our simplification experiments, we selected sentences from the test split of CEFR-SP whose readability annotations are publicly-available (i.e.\ from WikiAuto and SCoRE), and where both annotators agreed on their readability being above B1. 
This resulted in a total of 401 sentences.
As shown in Table~\ref{tab:test-dataset}, around 65\% of sentences in our test set belong to levels C1 and C2, indicating that models will need to deal with mostly ``diffcult'' input sentences.
Surprisingly, the readability levels show minimal variation in syntactic and lexical complexities, as indicated by measures of syntactic tree depth and age of acquisition, respectively.
However, sentences at the C2 level (the highest readability)  present the highest values in both indices.
\begin{table}[t]
    \centering
\small
    \begin{tabular}{@{}lcccc@{}}
    \toprule
         \textbf{Level}&  \textbf{Num. Sents.}&   \textbf{Avg. Len.} &\textbf{Avg. TD}&  \textbf{Avg. AoA}\\
    \midrule
         B2&  142&   18.43&6.38&  4.94\\
         C1&  187&   17.74&6.11&  4.95\\
         C2&  72&   18.87&6.65&  5.17\\
    \bottomrule
    \end{tabular}
    \caption{Test set statistics: the total number of sentences per level (Num. Sents), average sentence length (Avg Len), average tree depth (Avg TD) and average Kuperman age of acquisition score (Avg AoA).}
    \label{tab:test-dataset}
\end{table}

\subsection{Automatic Evaluation Metrics}
\label{sec:Automatic Evaluation Metrics}

We evaluated the generated simplifications using automatic metrics to check if the target readability was met and the original meaning preserved:

\paragraph{Adjacency CEFR Accuracy:} measures the percentage of sentences for which the system's output aligns closely with the target CEFR level. Specifically, it considers outputs successful if their CEFR level is within one level of the specified target. To predict CEFR levels, we use the state-of-the-art sentence-level estimator of \citep{arase-etal-2022-cefr}.\footnote{\url{https://zenodo.org/records/7234096}}  
    In preliminary experiments, LLMs and other models had lower performances when predicting CEFR levels (Appendix~\ref{app:CEFR-Level-Assessment-Performance}).

\paragraph{Root Mean Squared Error (RMSE):} measures the average error between the estimated and target CEFR levels of the system output, by calculating the square root of the average squared differences.
    
\paragraph{Spearman Correlation ($\rho$):} measures rank-order correlation between target and estimated CEFR levels of the system outputs.
    
\paragraph{Semantic Textual Similarity (STS):} measures the cosine similarity between the original and generated simplification embeddings. Higher scores reflect better meaning preservation in the output. We use the sentence transformer library \citep{reimers-2019-sentence-bert} to compute the sentence embeddings using the \textbf{all-MiniLM-L6-v2} model.

\paragraph{Exact Copies:} measures the percentage of generated simplifications that are exact matches to their corresponding original texts.   

\vspace{\baselineskip}
We also assessed fluency, but scores were high across all models and evaluation settings, so we only describe them in Appendix~\ref{app:Fluency-Evaluation}.

\begin{table*}[!ht]
\centering \small
\begin{tabular}{@{}lccc|cc@{}}
\toprule
Models & \textbf{$\rho$ (↑)} & \textbf{AdjAcc (\%) (↑)} & \textbf{RMSE (↓)} & \textbf{STS (\%) (↑)} & \textbf{Copies (\%) (↓)} \\
\midrule
\rowcolor{Gray}\multicolumn{6}{c}{\textit{Baselines}} \\
\midrule
\textbf{COPY} & 0 & 24.15 & 2.45 & 100 & 100 \\
\textbf{SUPERVISED} & 0.01±0.02 & 39.43±1.44 & 2.08±0.04 & 84.51±0.75 & 12.69±0.91 \\
\midrule
\rowcolor{Gray}\multicolumn{6}{c}{\textit{P1: target level}} \\
\midrule
Flan-T5-small & -0.01±0.02 & 34.69±0.76 & 2.21±0.01 & 86.70±0.58 & 12.97±1.12 \\
Flan-T5-base & -0.01±0.01 & 28.01±0.32 & 2.38±0.02 & 95.41±0.41 & 30.15±2.24 \\
Flan-T5-large & -0.00±0.01 & 28.04±0.49 & 2.36±0.01 & 96.40±0.25 & 25.68±1.58 \\
Flan-T5-xl & -0.00±0.01 & 27.96±0.89 & 2.38±0.01 & 97.17±0.20 & 30.09±1.39 \\
GPT-3.5-Turbo-Instruct & 0.05±0.01** & 39.87±1.03 & 2.01±0.03 & 85.12±0.36 & 0.00±0.00 \\
GPT-4-Turbo & \underline{0.11±0.01**} & \underline{51.68±0.79} & \underline{1.67±0.02} & 76.96±0.33 & 0.00±0.00 \\
OpenChat\_3.5 & -0.01±0.02 & 42.87±0.29 & 1.96±0.01 & 85.35±0.28 & 0.11±0.14 \\
Llama-3-8B-Instruct & 0.07±0.00** & 44.81±1.57 & 1.89±0.03 & 84.32±0.23 & 0.58±0.37 \\
Mistral-7B-Instruct-v0.2 & 0.06±0.01** & 41.40±0.83 & 1.98±0.01 & 83.63±0.20 & 0.00±0.00 \\
\midrule
\rowcolor{Gray}\multicolumn{6}{c}{\textit{P2: target level + description}} \\
\midrule
Flan-T5-small & 0.02±0.03 & 33.25±0.92 & 2.23±0.03 & 86.99±0.40 & 13.13±1.29 \\
Flan-T5-base & -0.01±0.01 & 27.60±0.70 & 2.39±0.02 & 95.54±0.45 & 29.54±2.27 \\
Flan-T5-large & 0.01±0.02 & 28.29±0.88 & 2.36±0.01 & 96.19±0.35 & 25.96±3.25 \\
Flan-T5-xl & 0.01±0.01 & 28.49±1.03 & 2.37±0.02 & 96.95±0.33 & 28.43±1.05 \\
GPT-3.5-Turbo-Instruct & 0.08±0.02** & 41.51±0.80 & 1.98±0.02 & 84.95±0.23 & 0.00±0.00 \\
GPT-4-Turbo & \underline{0.18±0.02**} & \underline{52.04±0.69} & \textbf{1.66±0.01} & 77.24±0.25 & 0.00±0.00 \\
OpenChat\_3.5 & 0.03±0.02 & 45.08±1.08 & 1.90±0.02 & 83.04±0.41 & 0.17±0.11 \\
Llama-3-8B-Instruct & 0.17±0.01** & 46.33±1.01 & 1.83±0.01 & 83.29±0.35 & 0.33±0.19 \\
Mistral-7B-Instruct-v0.2 & 0.09±0.01** & 41.65±0.88 & 1.96±0.01 & 83.45±0.31 & 0.00±0.00 \\
\midrule
\rowcolor{Gray}\multicolumn{6}{c}{\textit{P3: target level + examples}} \\
\midrule
Flan-T5-small & -0.00±0.01 & 38.79±0.89 & 2.12±0.02 & 78.54±0.49 & 6.48±1.28 \\
Flan-T5-base & -0.01±0.01 & 29.01±0.90 & 2.35±0.01 & 93.62±0.32 & 22.94±1.42 \\
Flan-T5-large & -0.01±0.00 & 28.43±0.40 & 2.36±0.01 & 95.75±0.17 & 23.30±3.17 \\
Flan-T5-xl & -0.01±0.01 & 28.10±0.55 & 2.37±0.01 & 96.86±0.26 & 31.78±1.35 \\
GPT-3.5-Turbo-Instruct & 0.08±0.00** & 42.12±0.67 & 1.98±0.01 & 84.97±0.55 & 0.00±0.00 \\
GPT-4-Turbo & \underline{0.15±0.00**} & \underline{50.73±1.17} & \underline{1.69±0.01} & 78.13±0.19 & 0.00±0.00 \\
OpenChat\_3.5 & 0.06±0.01* & 43.11±0.68 & 1.94±0.01 & 84.63±0.27 & 0.42±0.27 \\
Llama-3-8B-Instruct & 0.11±0.02** & 46.35±0.73 & 1.84±0.01 & 83.46±0.38 & 0.61±0.31 \\
Mistral-7B-Instruct-v0.2 & 0.08±0.01** & 40.23±0.64 & 2.03±0.01 & 85.66±0.28 & 0.00±0.00 \\
\midrule
\rowcolor{Gray}\multicolumn{6}{c}{\textit{P4: target level + description + examples}} \\
\midrule
Flan-T5-small & -0.02±0.00 & 38.51±1.02 & 2.13±0.01 & 77.83±0.70 & 7.70±1.08 \\
Flan-T5-base & -0.01±0.02 & 30.34±1.31 & 2.33±0.03 & 91.63±0.28 & 19.59±1.21 \\
Flan-T5-large & -0.01±0.01 & 29.01±0.53 & 2.35±0.01 & 95.51±0.27 & 22.58±1.71 \\
Flan-T5-xl & -0.01±0.00 & 28.70±0.69 & 2.36±0.02 & 96.57±0.24 & 28.90±1.84 \\
GPT-3.5-Turbo-Instruct & 0.08±0.01** & 43.34±1.15 & 1.93±0.02 & 84.40±0.23 & 0.00±0.00 \\
GPT-4-Turbo & 0.18±0.01** & \textbf{52.45±0.96} & \textbf{1.66±0.01} & 77.81±0.27 & 0.00±0.00 \\
OpenChat\_3.5 & 0.06±0.01* & 44.00±0.80 & 1.91±0.02 & 83.74±0.38 & 0.25±0.18 \\
Llama-3-8B-Instruct & \textbf{0.19±0.01**} & 48.21±1.23 & 1.80±0.02 & 82.86±0.16 & 0.50±0.19 \\
Mistral-7B-Instruct-v0.2 & 0.09±0.00** & 42.01±0.76 & 1.98±0.01 & 84.20±0.21 & 0.00±0.00 \\
\bottomrule
\end{tabular}
\caption{Results on the test set, across the four prompts. Metrics: Spearman's $\rho$, AdjAcc, RMSE, STS, and Copy Percentage. Boldface indicates the best overall performance across all prompts, while underlined results represent the best performance for each individual prompt. Significance: * (p < 0.01), ** (p < 0.001).}
\label{tab:result-by-prompts}
\end{table*}

\subsection{Baselines}

\paragraph{COPY:} System that simply copies the original text for each target readability level, setting the lowest standard for metric evaluation.
  
\paragraph{SUPERVISED:} Following \citet{scarton-specia-2018-learning}, we trained a level-based supervised baseline that incorporates a special token to indicate the target CEFR level (e.g.,\ <A1>, <A2>, <B1>) at the beginning of each original sentence. Due to the unavailability of original-simplified sentence pairs annotated with CEFR levels, we fine-tuned a pre-trained T5-Base \citep{2020t5} model on a specially curated unsupervised corpus.\footnote{Details about model training, training dataset creation, and inference settings are provided in Appendix~\ref{sec:appendixC}.}

\vspace{\baselineskip}

Table~\ref{tab:result-by-prompts} presents the scores obtained by all selected models in our test set, with the different designed prompts, and across all automatic metrics. The following sections analyze these results in terms of two core aspects: (1) the achievement of target readability levels (Sec.~\ref{sec:results-readability}), which ensures the simplified text meets intended accessibility goals, and (2) the preservation of the original meaning (Sec.~\ref{sec:results-meaning}), essential for maintaining fidelity to the source content. This dual focus is crucial for evaluating the effectiveness of the models in producing accessible, yet accurate, simplified texts.

\section{Are Target Readability Levels Met?}
\label{sec:results-readability}

\begin{figure*}[t]
    \centering
    \begin{subfigure}[b]{0.25\linewidth}
        \centering
        \includegraphics[width=\linewidth]{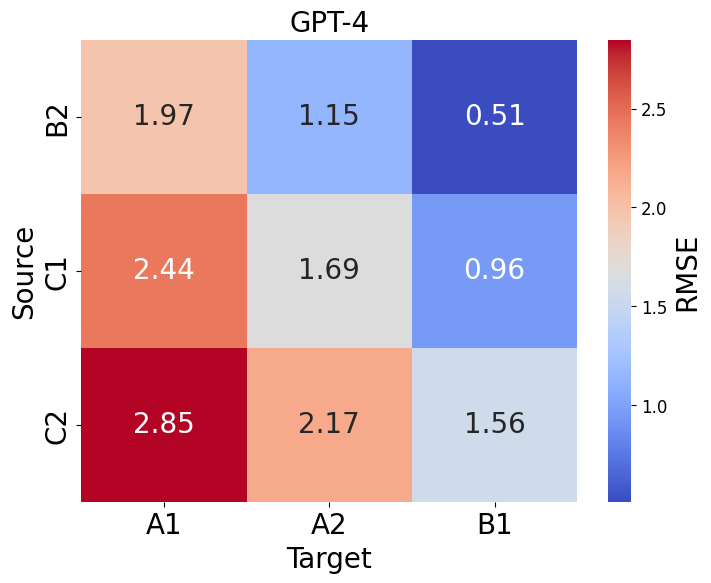}
        \label{fig:gpt4rmse}
    \end{subfigure}
    \begin{subfigure}[b]{0.25\linewidth}
        \centering
        \includegraphics[width=\linewidth]{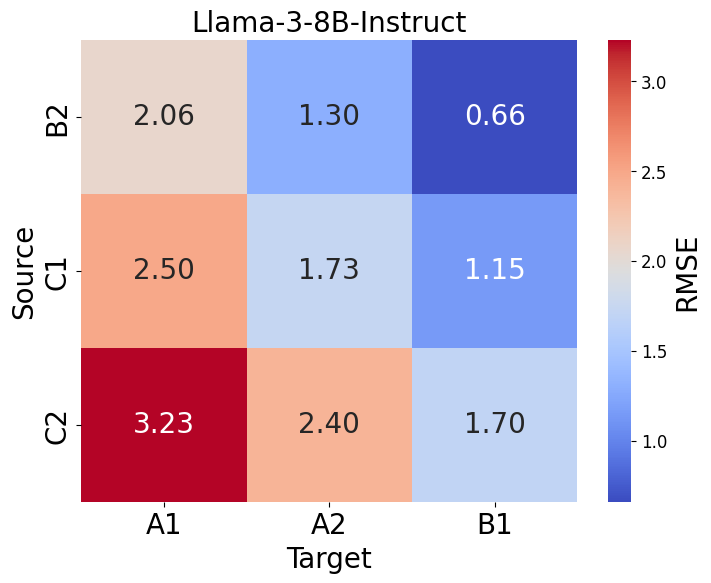}
        \label{fig:Llama-3-8B-Instructrmse}
    \end{subfigure}
    \begin{subfigure}[b]{0.25\linewidth}
        \centering
        \includegraphics[width=\linewidth]{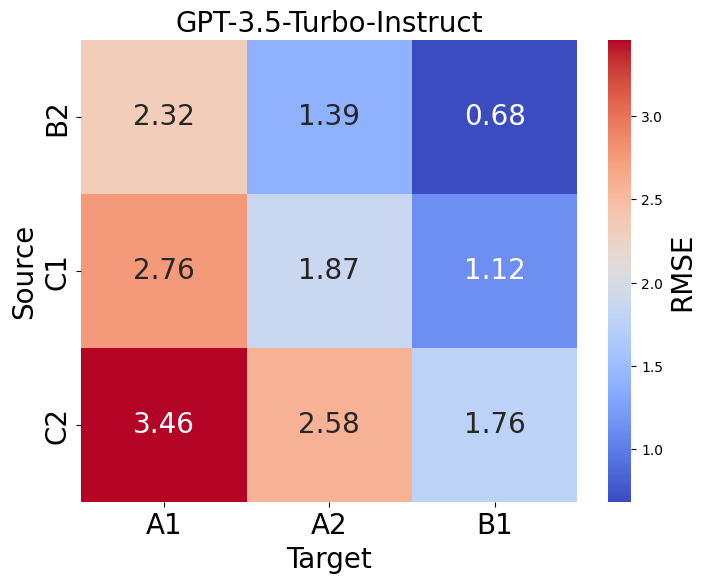}
        \label{fig:gpt3RMSE}
    \end{subfigure}
    \begin{subfigure}[b]{0.25\linewidth}
        \centering
        \includegraphics[width=\linewidth]{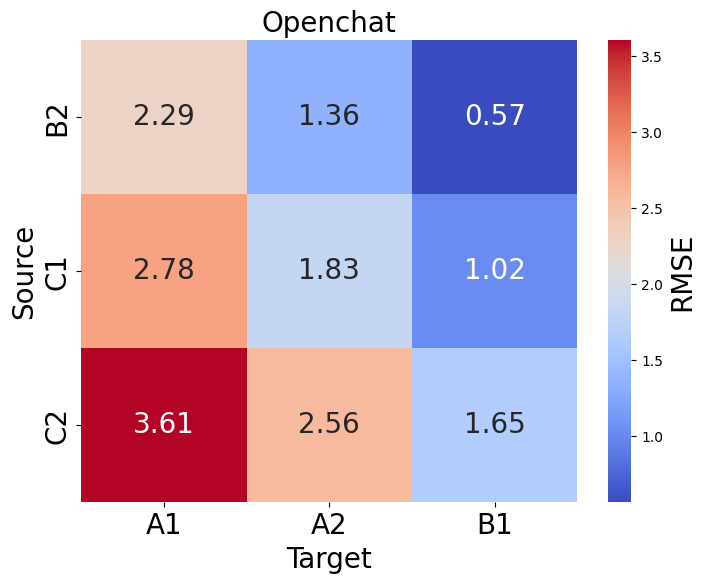}
        \label{fig:openchatRMSE}
    \end{subfigure}
    \begin{subfigure}[b]{0.25\linewidth}
        \centering
        \includegraphics[width=\linewidth]{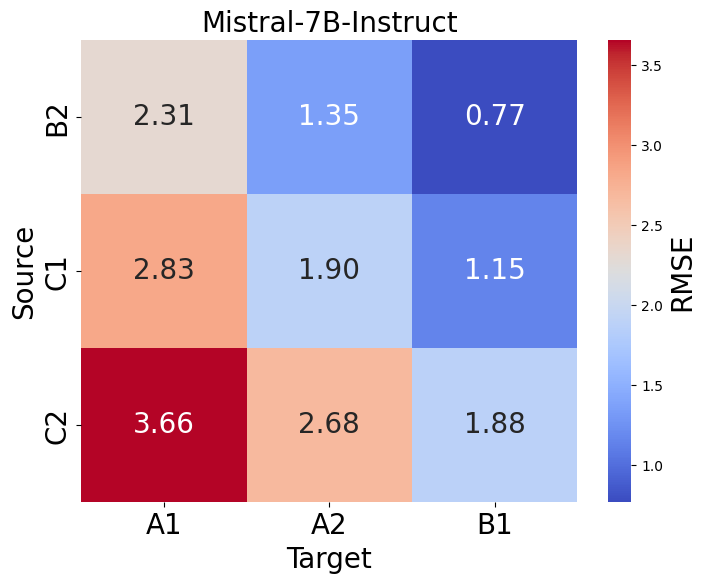}
        \label{fig:MistralRMS}
    \end{subfigure}
    
    \caption{Heatmaps depicting RMSE scores of the best models based on source and target readability levels.}
    \label{fig:combinedfig}
\end{figure*}

According to the scores in Table~\ref{tab:result-by-prompts} (left side), all models have difficulties in reaching the target CEFR levels, independently of the prompt being used.
GPT-4 with P4 emerges as the top-performing model, but it only achieves a  CEFR accuracy of 52.45\% and a low Spearman’s rank correlation of 0.18. Its RMSE of 1.66 suggests that the model tends to generate simplifications with more than one CEFR level difference from the target.
Llama-3-8B-Instruct also shows a similar performance, with accuracies ranging from 44.81\% to 48.21\%, $\rho$ values between 0.07 and 0.19, and RMSE scores are close to 2.

Compared to the SUPERVISED baseline (accuracy of 39.43\%, near-zero correlation, and RMSE of 2.08), the previous two models show better readability control. 
Other models, such as Openchat, GPT-3.5-turbo-instruct and Mistral-7B-Instruct-v0.2, show moderate improvements over SUPERVISED, with accuracies between 40.23\% and 45.08\%, low positive correlations, and RMSE values ranging from 1.90 to 2.03. 
In contrast, Flan-T5 models demonstrate poor performance in the task, under-performing even against the SUPERVISED baseline with accuracies ranging from 27.60\% to 38.79\%, mostly negative or negligible correlations, and high RMSE values (between 2.12 and 2.39).
Actually, these models have performances close to the COPY baseline, indicating a conservative behavior.
The following sections provide further insights into these results.

\subsection{The Effect of Target CEFR Information}
Providing information about the target readability, in the form of  level descriptions and example sentences, resulted in a slight enhancement of performance. 
This is evidenced by the increase in accuracy and decrease in RMSE when comparing prompts P1 to P2, P3 and P4. 
Most models achieved their best performance when combining both CEFR descriptions and example sentences (P4). 
However, Openchat\_3.5 exhibited greater improvements from descriptions alone rather than from examples, attaining its highest accuracy and lowest RMSE with prompt P2. 
Similarly, the Mistral-7B-Instruct-v0.2 model showed a positive response to descriptions, with better performance in P2 compared to P1 and P3, and reached its peak accuracy with the full information provided in P4. 
In contrast, the Flan-T5 small and base models experienced a decline in performance when provided exclusively with CEFR descriptions (P2), performing slightly worse than with the base prompt (P1).
Overall, responses to additional contextual information about target readability varies across models.

\begin{figure*}[t]
    \centering
  
    \begin{subfigure}[b]{0.48\linewidth}
        \centering
        \includegraphics[width=\linewidth]{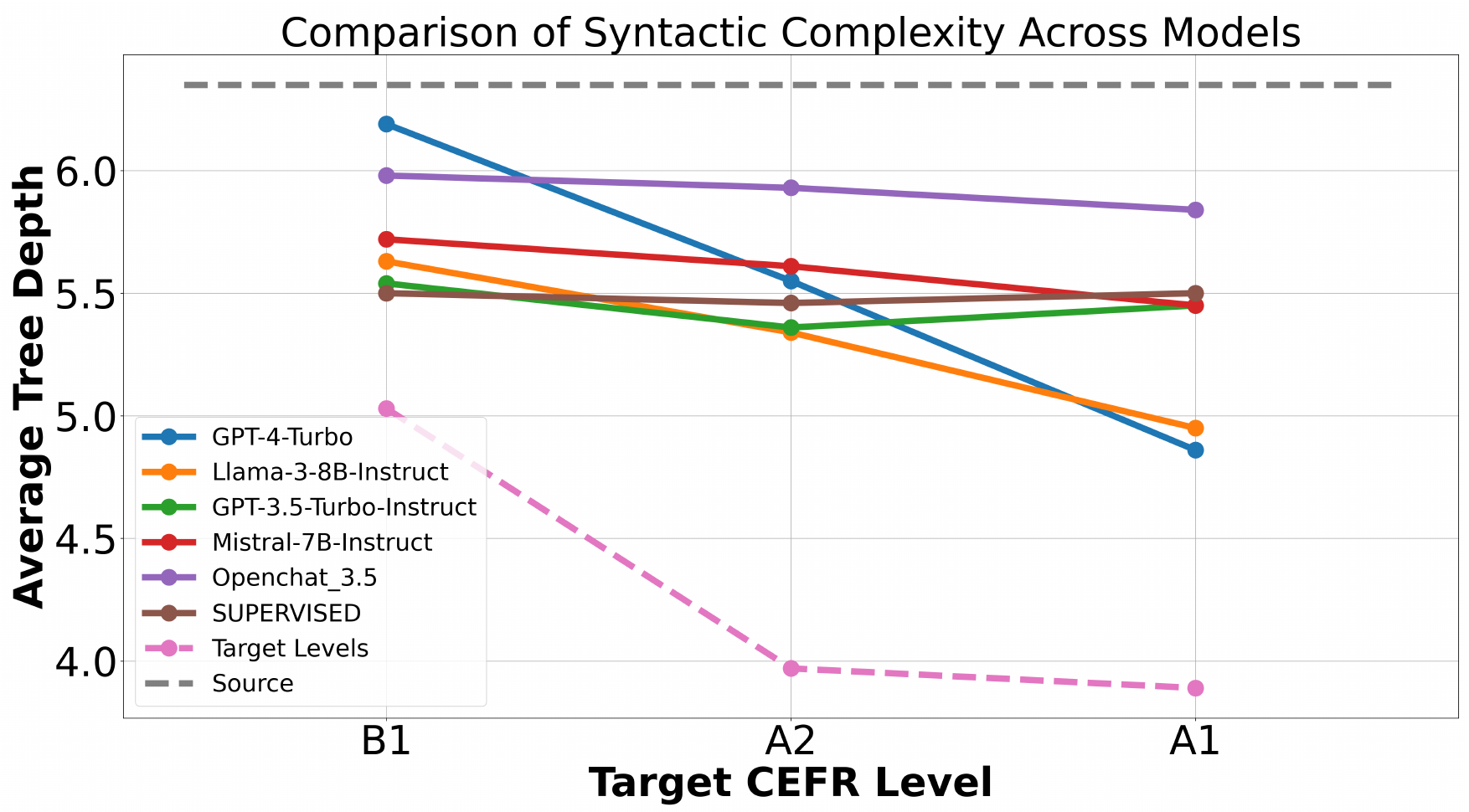}

        \label{fig:sydiff}
    \end{subfigure}
    \hfill
    \begin{subfigure}[b]{0.48\linewidth}
        \centering
        \includegraphics[width=\linewidth]{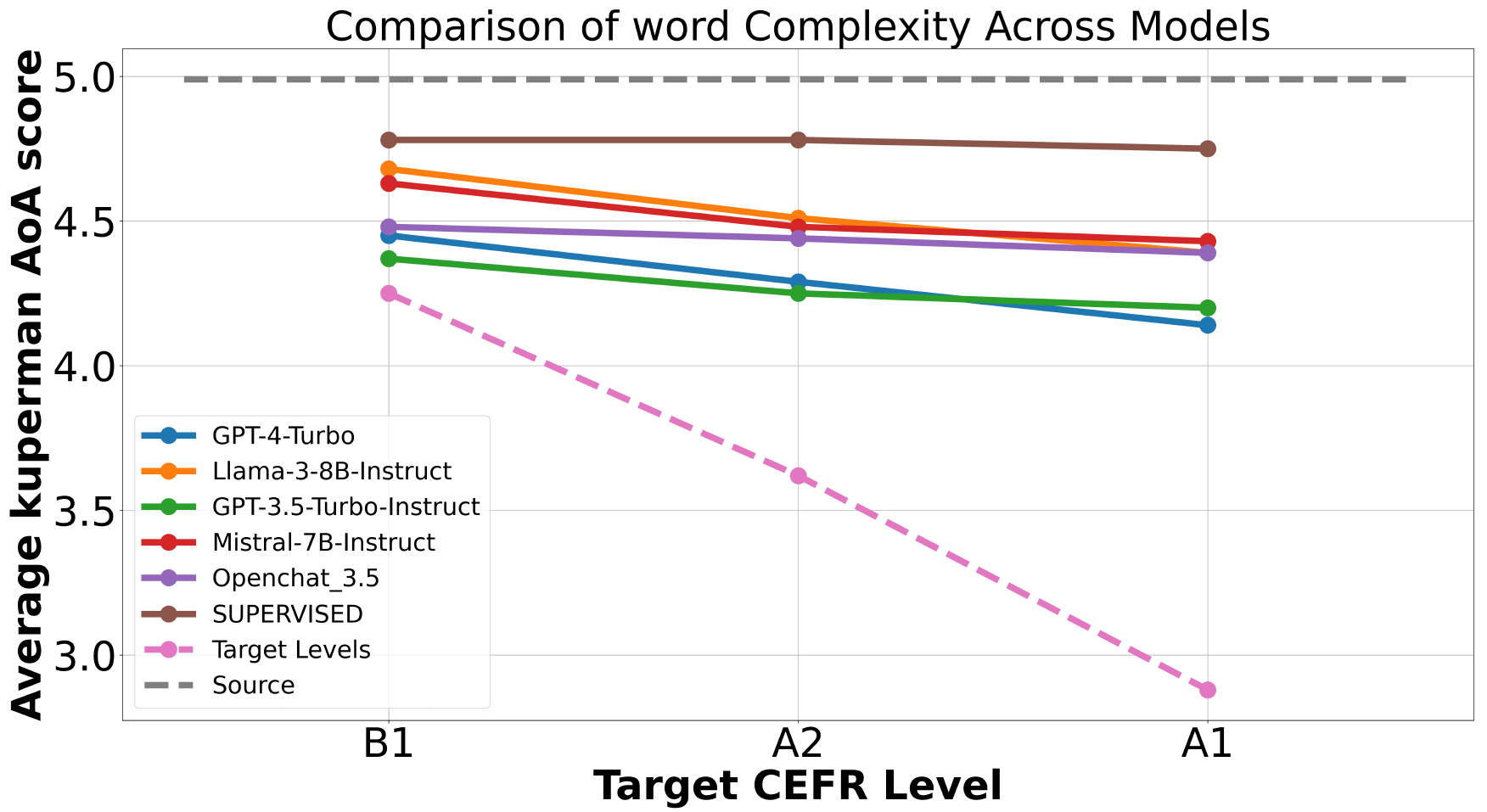}

        \label{fig:lexdiff}
    \end{subfigure}
   
    \caption{Line graphs comparing syntactic and lexical complexities of the generated simplifications from the top-performing models to the source and target level syntactic and lexical complexity.}
    \label{fig:lex-sy-diff}
\end{figure*}

\subsection{The Effect of Source vs Target Readability}
\label{sec:Source vs Target}

Figure~\ref{fig:combinedfig} presents the RMSE scores for the top five performing models across all combinations of source and target CEFR levels. 
The heatmaps indicate that simplification becomes increasingly difficult (evidenced by higher RMSE) as target levels decrease. 
Models achieve their best performance when source and target CEFR levels are close, especially in simplifications from B2 to B1, as shown by the blue sections.
This suggests that models handle moderate simplifications with fewer complexity adjustments more effectively.

Conversely, models struggle most as the gap between source and target levels widens, especially when simplifying from C2 to A1, highlighting their limited ability to handle large reductions in complexity.

\subsection{Are Generated Outputs ``Simple''?} 

We analyse if the generated outputs are still ``simple'', even if they did not meet their target readability levels.
In particular, we look at the average syntactic tree depth and the average Kuperman Age of Acquisition (AoA) score,\footnote{Calculated with LFTK \citep{lee-lee-2023-lftk}.} as proxies for syntactic and lexical complexity, respectively.
We compute these scores considering all original sentences, and all generated simplifications in each target CEFR level.
As a reference, we also compute them for all sentences in the CEFR-SP dataset manually annotated with the target readability levels (B1, A2 an A1).
Figure~\ref{fig:lex-sy-diff} compares these scores for the simplifications of the top five models, the SUPERVISED baseline, and both source and target references.

Regarding syntactic complexity, simplifications from the SUPERVISED baseline exhibit minimal variation across all target CEFR levels, suggesting limited adaptability. 
In contrast, GPT-4-Turbo and Llama-3-8B-Instruct adjust syntactic complexity more effectively as the target CEFR level decreases.
Regarding lexical complexity, most LLMs reduced complexity at lower CEFR levels. 
The SUPERVISED baseline's lexical complexity remained relatively flat, again indicating a lack of adaptability to varying levels of proficiency. Notably, GPT-4 and GPT-3.5-Turbo-Instruct were particularly effective at adapting lexical complexity.
However, comparing these scores to the ones from the target levels, reveals significant room for improvement.

\subsection{Do Humans Agree with the Metrics?}

\begin{table}[tb]
    \centering
    \scriptsize
    \begin{tabular}{@{}lccc@{}}
     \toprule
         Model& Levels&  AccuracyH(\%)&  AccuracyA(\%) \\
         \midrule

& B1    &100.00&60.00 \\
GPT-4-Turbo& A2    &86.67&46.67 \\
&A1    &53.33&20.00\\
&Overall &  80&  42.2 \\
         \midrule

&B1    &93.33& 60.00\\
Llama-3-8B-Instruct&A2    &93.33&66.67 \\
&A1    &66.67& 6.67 \\
&Overall &  84.4&  44.4\\
         \bottomrule
     \end{tabular}
    \caption{Human readability assessment results for GPT-4-Turbo and Llama-3-8B-Instruct. AccuracyH: Human-rated accuracy; AccuracyA: Classifier-rated accuracy.}
    \label{tab:readability-assessment-Human}
\end{table}

We randomly sampled 45 simplifications (15 per target level) from  GPT-4-Turbo and Llama-3-8B-Instruct, totaling 90 instances. 
We recruited three evaluators with experience in teaching English as a second language, and asked them to: (1) assess the CEFR level that a student would need to understand each generated simplification, and (2) judge whether the corresponding source sentences could actually be adjusted to match the target CEFR levels. 
This aimed to establish if the fact that an output did not reach the target level was due to issues with the model itself or the nature of the source text.\footnote{Details on evaluator selection, inter-rater agreement, and the exact instructions provided can be found in Appendix~\ref{app:Human-Readability-Assessment-Details}.} 

Once again, we compute Adjacency CEFR Accuracy, but this time using the manual judgements instead of predicted levels. The majority label was set as the CEFR level for each instance. If no clear majority existed, we checked if the annotated levels differed by more than one from the pre-specified target. If so, the simplification was considered incorrect; otherwise, it was considered correct.

\paragraph{Models mostly reach the CEFR targets.}
Table~\ref{tab:readability-assessment-Human} shows that both GPT-4-Turbo and Llama-3-8B-Instruct produced sentences that aligned with the target CEFR levels in 80\% and 84.4\% of cases, respectively, based on human evaluations (AccuracyH). This indicates that in the majority of cases, human raters judged the generated sentences to match closely the expected readability levels for the intended audience. 

\paragraph{Models struggle with lower CEFR levels.}
Both models struggled when simplifying sentences to lower CEFR levels, particularly A1, confirming the finding in~\ref{sec:Source vs Target}. GPT-4 failed in 7 A1 cases and 2 A2 cases out of 9 total failures, while Llama 3 failed in 4 A1 cases and 1 each for A2 and B1 out of 6 total failures. 
However, for failed cases, especially at A1 level, evaluators judged most source sentences unsuitable for adjustment: GPT-4 had 6 out of 7 A1 cases, and Llama 3 had 3 out of 4 A1 cases judged as nonadjustable. Overall, while both models excelled at higher CEFR levels, simplifying to lower levels remains challenging due to both model limitations and the inherent constraints of limited language proficiency at lower readability levels.

\begin{table}[tb]
\centering
\scriptsize
\begin{tabular}{@{}p{4cm}p{1.2cm}p{1.2cm}@{}}
\toprule
\textbf{Sentence} & \textbf{RatingH} & \textbf{RatingA} \\
\midrule
Primary hypersomnias, which are conditions that make people very sleepy, are not common.  & B1 & C1 \\
\midrule
When a woman is giving birth, the water bag (amniotic sac) breaks and the water (amniotic fluid) comes out from her body. & A2 & C1 \\
\midrule
The students were happy because the Japanese Women's Team won. They stayed awake late. They cleaned up before they went home. & A2 & B1 \\
\bottomrule
\end{tabular}
\caption{Comparison of Human and Classifier Ratings}
\label{tab:HvsA}
\end{table}

\paragraph{Automatic CEFR prediction does not account for all types of simplification operations.}
To understand the gap between AccuracyH and AccuracyA, we reviewed some instances that show differences between classifier and human evaluations. 
For the first example in Table~\ref{tab:HvsA}, human evaluators rated it as B1, while the classifier rated it as C1. However, if we removed the technical term ''Primary hypersomnias'', the classifier re-rates the sentence as B1. This shows that the classifier tends to assign a higher level when technical terms are present, even if they are explained in the sentence itself. 
Similarly, for the second example of Table~\ref{tab:HvsA}, human evaluators rated it as A2, while the classifier rated it as C1. When the medical terms ``amniotic sac'' and ``amniotic fluid'' are removed, the classifier revised the rating to B1, again indicating that it overestimates complexity when presented with technical terms, even if explanations are provided. 
For the third example, human evaluators rated it as A2, while the classifier rated it as B1. However, when each sentence is evaluated individually, the classifier assigns ratings of A2, A1, and A2, respectively, aligning more closely with human judgment. 

This highlights key areas where the classifier’s approach differs from human judgment, and suggests the need for further improvement.  Elaboration and sentence splitting are vital simplification operations, particularly at lower CEFR levels, and failing to account for them leads to inaccurate difficulty assessments. Training the classifier with parallel sentences—where the same text is written at different CEFR levels (e.g., A1, A2, B1) could help the model better understand sentence complexity and align more closely with human evaluations.

\section{Is the Original Meaning Preserved?}
\label{sec:results-meaning}

\begin{table}[tb]
\centering
\scriptsize
\begin{tabular}{@{}p{2.9cm}p{3.5cm}c@{}}
\toprule
\textbf{Original} & \textbf{Simplified} & \textbf{STS} \\
\midrule
(\textbf{C2}) Pharmacoepidemiology bridges pharmacology and epidemiology and borrows their methodology from both of them. & (\textbf{B1}) (\textbf{GPT-4})  Pharmacoepidemiology combines the study of drugs (pharmacology) and the study of health in populations (epidemiology). It uses methods from both these fields. & 0.82 \\
\midrule
(\textbf{B2}) Nothing you do or say will ever stop Sally from chasing her dream of becoming a veterinarian. & (\textbf{A1}) (\textbf{Llama3}) Sally will always try to become a vet. & 0.78 \\
\midrule
(\textbf{C1}) There are two main kinds of charter: bareboat and skippered. & (\textbf{B1}) (\textbf{GPT-4}) 
There are two main types of boat rental: one where you rent the boat without a crew (bareboat), and one where you rent the boat with a captain (skippered). & 0.60 \\

\midrule
(\textbf{C1}) However, some people with schizophrenia have the ability to tickle themselves. & (\textbf{B1}) (\textbf{GPT-4}) Some people who have schizophrenia can tickle themselves. & 0.97 \\
\bottomrule
\end{tabular}
\caption{Examples of the impact of simplification on STS scores.}
\label{tab:examples}
\end{table}

\subsection{Vector-based Semantic Similarity}

According to the scores in Table~\ref{tab:result-by-prompts} (right side), models show varied performances. 
Flan-T5 models (particularly Flan-T5-large and Flan-T5-xl) achieve the highest STS scores, ranging from 95.51\% to 96.95\%. 
This superior performance is likely attributed to their conservative behavior (high percentage of Copies). 
Other models such as Llama-3-8B-Instruct, Openchat, Mistral-7B-Instruct-v0.2, and GPT-3.5-turbo-instruct achieve scores ranging from 83.04\% to 85.66\%. 
However, GPT-4, while excelling in readability control, often shows lower STS scores, ranging from 76.96\% to 78.13\%. 

Through manual inspection, we found that STS tends to over-penalize text changes that could be considered appropriate in RCTS. 
In order to make a text more accessible, simplification can sometimes involve removing or inserting new information, such when executing compression or elaboration operations, respectively \citep{alva-manchego-etal-2020-data}. 
Such forms of rewriting might result in STS scores that underestimate the true quality of the simplifications.
For example, the first three simplifications in Table \ref{tab:examples} could be considered of high quality since they accurately capture the meaning of their original counterparts. 
However, they did not receive the nearly-perfect STS scores that the last conservative simplification did.
Experiments with BERTscore \citep{bert-score} and AlignScore \citep{zha-etal-2023-alignscore}, typically used for meaning preservation and factuality evaluation, yielded similar results.
As shown in Table~\ref{tab:bertscoreexamples}, BERTscore favored outputs that minimally modify or directly copy the input text.
On the other hand, AlignScore penalized the addition of supporting details (such as elaboration and examples) deeming them incorrect (as illustrated in the first and second examples in Table~\ref{tab:examplesAlignScore}), while rewarding factual accuracy even when simplifications are inappropriate (as seen in the third example in Table~\ref{tab:examplesAlignScore}).\footnote{More details and results for BERTscore and AlignScore are available in Appendix~\ref{app:bertscore-alignscore-results}}
This highlights the necessity for meaning preservation metrics that account for the nuances of the editing operations executed while simplifying texts.

\begin{table}[tb]
\centering
\scriptsize
\begin{tabular}{@{}p{2.4cm}p{3.2cm}c@{}}
\toprule
\textbf{Original} & \textbf{Simplified} & \textbf{BERTscore} \\
\midrule
(\textbf{C2}) Pharmacoepidemiology bridges pharmacology and epidemiology and borrows their methodology from both of them. & (\textbf{B1}) (\textbf{GPT-4})  Pharmacoepidemiology combines the study of drugs (pharmacology) and the study of health in populations (epidemiology). It uses methods from both these fields. & 0.52 \\
\midrule

(\textbf{C1}) Rivlin had pulmonary fibrosis, a lung disease. & (\textbf{B1}) (\textbf{GPT-4}) 
Rivlin had a lung disease called pulmonary fibrosis. & 0.86 \\
\midrule

(\textbf{C2}) Dimethylmercury is extremely toxic and dangerous to handle. & (\textbf{B1}) (\textbf{Mistral-7B}) Dimethylmercury is very harmful and should be handled with care. & 0.75 \\
\midrule
(\textbf{C2}) Dimethylmercury is extremely toxic and dangerous to handle.& (\textbf{B1}) (\textbf{Flan-T5-L}) Dimethylmercury is extremely dangerous and toxic to handle. & 0.95 \\
\bottomrule
\end{tabular}
\caption{Examples of the impact of simplification on BERTscore scores.}
\label{tab:bertscoreexamples}
\end{table}

\begin{table}[t]

 \centering

 \scriptsize

 \begin{tabular}{@{}p{2.5cm}p{3.2cm}c@{}}

 \toprule

 \textbf{Original} & \textbf{Simplified} & \textbf{AlignScore} \\
 \midrule

 (\textbf{C1}) Primary hypersomnias are very rare. & (\textbf{A1}) (\textbf{GPT-4}) Hypersomnias are sleep conditions where someone sleeps a lot or feels very sleepy during the day. Primary hypersomnias are a type of this condition and they do not happen often. They are rare. & 0.003 \\

 \midrule

 (\textbf{C1}) Bisques are heavy cream soups traditionally prepared with shellfish, but can be made with any type of seafood or other base ingredients. & (\textbf{A1}) (\textbf{GPT-4}) Bisques are thick soups with cream. They are usually made with shellfish like crab or lobster. But you can use other seafood or different ingredients too. & 0.44 \\

 \midrule

 (\textbf{B2}) Unfortunately, she is a parent whose rules must be enforced through guilt and manipulation instead of reason and trust. & (\textbf{A1}) (\textbf{Flan-T5-small}) This is a parent. & 0.97 \\

 \bottomrule

 \end{tabular}

 \caption{Examples illustrating the impact of simplification on AlignScore scores.}

 \label{tab:examplesAlignScore}

 \end{table}

\subsection{Human Judgements}
We randomly sampled 45 simplifications from GPT-4-Turbo and Llama-3-8B-Instruct, totaling 90 instances. 
Three fluent English speakers evaluated whether the original meaning was retained in the simplified sentences. 
We adapted the error annotation scheme for evaluating factuality in TS proposed by \citet{devaraj-etal-2022-evaluating}: instead of scoring each operation independently, we asked annotators to provide an overall score for each simplification.
Annotators rated each simplification using the following scale: 0 indicates \textit{no or trivial changes}; 1 signifies that \textit{non-trivial changes have occurred, but the main idea is maintained}; and 2 indicates that \textit{the main idea has not been preserved in the simplified sentence}. 
We also allowed intermediate scores (0.5 and 1.5).\footnote{Details on evaluator selection, inter-rater agreement, and the instructions provided can be found in Appendix~\ref{app:Human-Meaning-Evaluation-Details}.} 
\begin{table}[tb]
\centering
\setlength{\tabcolsep}{2pt} 
\scriptsize
\begin{tabular}{@{}l|c|cccccc@{}}

\toprule

Model& STS(\%) &   Avg &  0 (\%) &   0.5 (\%) &   1 (\%) &  1.5 (\%) &  2 (\%)\\

\midrule

GPT-4-Turbo & 67.41    & 0.94& 8.88&13.33 &62.22&  11.85& 3.70\\
\midrule

Llama-3-8B-Instruct& 79.92    &  0.95& 17.03& 22.96 &25.92
&20.0&   14.07\\

\bottomrule
\end{tabular}
\caption{Human evaluation of meaning preservation.}
\label{tab:human-MP-Results}
\end{table}


As shown in Table~\ref{tab:human-MP-Results}, Llama-3 obtained a higher percentage of minimal (score 0.5) or no changes (score 0) compared to GPT-4-Turbo, which explains its higher STS score. However, Llama-3 also had more instances where meaning was significantly altered or lost (14.07\% scored 2) compared to GPT-4-Turbo (3.70\%). This contrasts with the STS results, which showed GPT-4-Turbo having the lowest meaning preservation. In contrast, GPT-4-Turbo made more non-trivial changes while still preserving the original meaning, with 62.22\% of its simplifications receiving a score of 1, which may explain its lower STS rating.
This discrepancy between STS scores and human evaluation underscores the challenge of assessing meaning preservation in simplification tasks, highlighting the need for better metrics that balance meaning retention and substantial rewriting.


\section{Related Work}
\paragraph{Readability-Controlled Simplification.}
To control the readability of simplified text, two main approaches have been proposed. The first modifies the models' training process by (a) adding special tokens \citep{scarton-specia-2018-learning,nishihara-etal-2019-controllable,spring-etal-2021-exploring,rios-etal-2021-new,yanamoto-etal-2022-controllable,zetsu-etal-2022-lexically,agrawal-carpuat-2023-controlling}) or (b) instruction prompts \citep{chi-etal-2023-learning} to the source sentences during training to guide the model's behavior. The second approach employs a post-processing control module to control readability during  inference \citep{zetsu-etal-2022-lexically}.
All these methods rely on fully supervised training approaches, requiring a large amount of sentence pairs annotated with readability levels. 
However, a major challenge lies in the difficulty of obtaining such annotated data. 
Our work sought to reduce dependence on parallel data by leveraging instruction-tuned LLMs for sentence-level RCTS in zero-shot settings, only relying on contextual information for each target readability level.

\paragraph{Text Simplification with LLMs.}

Few studies exist on LLMs' ability to perform TS
\citep{feng2023sentence,ryan-etal-2023-revisiting,kew-etal-2023-bless}, with limited exploration of their ability for controlling the readability of simplifications.  
\citet{imperial-tayyar-madabushi-2023-flesch} examined various instruction-tuned LLMs in interpreting and applying readability specifications from prompts in their output for story completion and narrative simplification. 
However, their evaluation was limited to one readability level and focused on readability achievement. 
Also, \citet{farajidizaji2023possible} investigated the potential of ChatGPT and Llama-2 for modifying passages for different readability levels.
Different from these works, our study included a higher number and range of LLMs, and focused on sentence-level RCTS rather than passage-level.
Furthermore, we not only assessed models on their ability to meet the target readability, but analysed how this affects meaning preservation using both automatic metrics and human assessment.


\section{Conclusion}

We explored sentence-level zero-shot RCTS to analyse various instruction-tuned LLMs in simplifying sentences to specified CEFR levels.
Our experiments showed that LLMs have difficulties with this task, especially when there is a large gap between source and target readability levels. 
Providing contextual information about the target readability (i.e.\ CEFR descriptions and example sentences) improves performance but not uniformly.

We also demonstrated that automatically assessing models for RCTS is challenging due to the nature of the task. 
Standard similarity metrics penalise removing or adding information even if it improves readability, while CEFR predictors often misjudge the true difficulty of simplified texts, failing to account for rewritings via elaboration and sentence splitting. 
This gap between human and automated assessments underscores the need for more nuanced evaluation mechanisms in RCTS. 
Refined metrics would not only enhance the evaluation process but also drive significant improvements in RCTS development by better supporting output filtering to ensure both readability and meaning preservation, or by serving as reward signals in reinforcement learning approaches.

\section*{Limitations}

Due to our focus on sentence-level RCTS, our study was constrained to the data in the CEFR-SP dataset, the only available resource with sentence-level readability annotations from experts.
As such, our experiments were exclusively conducted using English data and limited to the CEFR standard for establishing readability levels.

Additionally, our study focused on zero-shot scenarios.
While obtaining a large parallel dataset is challenging, collecting a few (synthetic) samples might be feasible, allowing for studies on few-shot or small-scale fine-tuning setting.
This is considered as future work.

\section*{Ethics Statement}
For our experiments, we only used subsets of the CEFR-SP dataset with non-restrictive licenses.
In particular, the Wiki-Auto dataset is publicly available under the CC BY-SA 3.0 license, and the SCoRE dataset is available under the CC BY-NC-SA 4.0 license.

For readability assessments, we paid the evaluators \$65.99 for completing the task, which took an average of 1 hour to complete. Similarly, for meaning preservation assessments, evaluators received a flat rate of \$33 to complete the task, with an average task completion time of 40 minutes.

\bibliography{anthology,custom}

\appendix

\begin{table*}[t]
\centering \small

\begin{tabular}{@{}c|m{400pt}@{}} 
\toprule
\textbf{CEFR levels} & \textbf{Descriptors} \\ 
\midrule
A1 & Can understand very short, simple texts a single phrase at a time, picking up familiar names, words and basic phrases and rereading as required. \\ \midrule 
A2 & Can understand short, simple texts containing the highest frequency vocabulary, including a proportion of shared international vocabulary items. \\  \midrule
B1 & Can read straightforward factual texts on subjects related to their field of interest with a satisfactory level of comprehension.\\ \midrule
B2 & Can read with a large degree of independence, adapting style and speed of reading to different texts and purposes, and using appropriate reference sources selectively. Has a broad active reading vocabulary, but may experience some difficulty with low-frequency idioms.\\ \midrule
C1 & Can understand a wide variety of texts including literary writings, newspaper or magazine articles, and specialised academic or professional publications, provided there are opportunities for rereading and they have access to reference tools.\\ & Can understand in detail lengthy, complex texts, whether or not these relate to their own area of speciality, provided they can reread difficult sections. \\ \midrule
C2 & Can understand a wide range of long and complex texts, appreciating subtle distinctions of style and implicit as well as explicit meaning.\\ & Can understand virtually all types of texts including abstract, structurally complex, or highly colloquial literary and non-literary writings. \\ 
\bottomrule
\end{tabular}
\caption{CEFR descriptors, describing readers' overall comprehension abilities for each CEFR level.}
\label{tab:cefr-descriptor}
\end{table*}

\begin{figure*}[t]
    \centering
    \includegraphics[width=\linewidth]{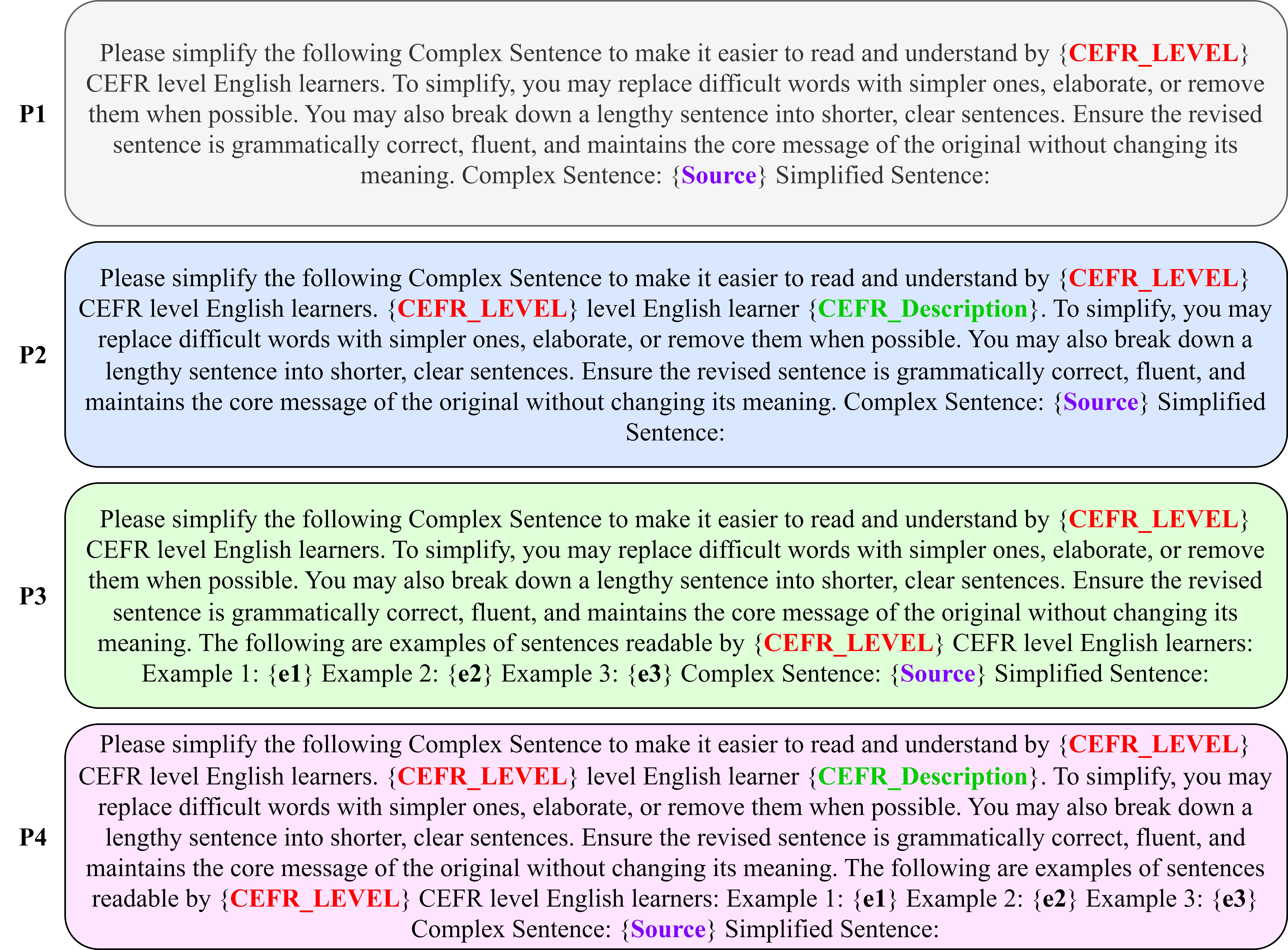}
    \caption{Prompts provided to instruction-tuned LLMs.}
    \label{fig:Prompts}
\end{figure*}

\section{Selected Instruction-Tuned LLMs}
\label{app:models-descriptions}
We investigate 8 instruction-tuned LLMs with diverse architectures and sizes, including open-weight and closed-weight models. Taking into account hardware resource limitations, we select open-source models within the maximum limit of 8 billion parameters.

\paragraph{Flan-T5} \citep{chung2022scaling} is an advanced open-source instruction-fine-tuned language model built on top of T5 \citep{2020t5}. In our study, we used four Flan-T5 variants: Small\footnote{\url{https://huggingface.co/google/flan-t5-small}}, Base\footnote{\url{https://huggingface.co/google/flan-t5-base}}, Large\footnote{\url{https://huggingface.co/google/flan-t5-large}}, and XL\footnote{\url{https://huggingface.co/google/flan-t5-xl}} with 80M, 250M, 780M, and 3B parameters, respectively. These models were fine-tuned across 1,836 fine-tuning tasks from instruction datasets including T0-SF \citep{sanh2022multitask}, Muffin \citep{wei2022finetuned}, and Natural Instructions V2 datasets \citep{wang-etal-2022-super}.

\paragraph{Mistral-7B-Instruct} is another open-source instruction-fine-tuned LLM created by fine-tuning the Mistral 7 billion parameters model \citep{jiang2023mistral} on various publicly shared instruction datasets in HuggingFace. We use the improved Instruct version 0.2 model.\footnote{\url{https://huggingface.co/mistralai/Mistral-7B-Instruct-v0.2}}

\paragraph{OpenChat-7B} \citep{wang2023openchat} is a collection of open-source language models, fine-tuned with the Conditioned-RLFT (C-RLFT) approach using mixed-quality data. For this study, we used the OpenChat-3.5 7B parameter model built on top of Mistral-7B.\footnote{\url{https://huggingface.co/openchat/openchat_3.5}}

\paragraph{Llama-3-instruct-8B} \citep{llama3modelcard} represents the latest advancement in Meta's instruction-tuned language models series. For this task, we use the 8B version finetuned using supervised fine-tuning (SFT) and reinforcement learning with human feedback (RLHF) on publicly available instruction datasets, along with over 10 million human-annotated training data.\footnote{\url{https://huggingface.co/meta-llama/Meta-Llama-3-8B-Instruct}}

\paragraph{OpenAI-GPT} in this study we use GPT-3.5-turbo and GPT-4-turbo from OpenAI. For GPT-3.5-turbo we used the version fine-tuned with proprietary training data up to Sep 2021 with a 4K context window.\footnote{gpt-3.5-turbo-instruct: \url{https://platform.openai.com/docs/models/gpt-3-5-turbo}} For GPT-4-turbo we used the version fine-tuned with proprietary training data up to April 2023 with a 128K context window.\footnote{gpt-4-1106-preview: \url{https://platform.openai.com/docs/models/gpt-4-turbo-and-gpt-4}}

\section{CEFR Levels Descriptors}
\label{app:CEFRLevelsDescriptors}

Table \ref{tab:cefr-descriptor} outlines the descriptors for each CEFR level, based on the overall reading comprehension scale used in the communicative language activities of the CEFR Descriptor Scheme.\footnote{\url{https://www.coe.int/en/web/common-european-framework-reference-languages/cefr-descriptors}}

\section{Prompts}
\label{app:appendix-prompts}
Figure~\ref{fig:Prompts} illustrates the four prompts provided for LLMs RCTS.

\section{Supervised Baseline Training and Inference Settings }
\label{sec:appendixC}
\subsection{Supervised Baseline Training Dataset}
This corpus was generated by applying the CEFR classifier \citep{arase-etal-2022-cefr} to the sentence alignments from the OneStopEnglish \citep{vajjala-lucic-2018-onestopenglish} dataset and Wiki-Manual (aligned, partialAligned) splits \citep{jiang-etal-2020-neural}. Then we select only sentence pairs with different CEFR levels and swap their source-target roles when the target sentence has a higher CEFR level. Additionally, we ensured there was no overlap between sentences used in testing and those extracted from the Wiki-Manual. This process resulted in the selection of 2557 sentence pairs --- 1,255 from OneStopEnglish and 1,302 from the Wiki-Manual. 20.94\% of sentence Alignments from OneStopEnglish and 44.75\% of sentence alignments from Wiki-Manual were removed because the source-simple pairs had been classified at the same CEFR level. Moreover, one sentence alignment was removed from Wiki-Manual as it overlapped with the test set. Then we split the dataset into 80\% training and 20\% validation sets.
Table \ref{tab:unsupervised_dataset_cefr_sample_counts_combined} provides the distribution of sentence target levels in both data splits.

\begin{table}[tb]
\centering
\small

\begin{tabular}{lrrrrrrr}
\toprule
{} &   A1 &   A2 &   B1 &   B2 &  C1 &  Total\\
\midrule
Train      &   96 &  473 & 1264 &  202 &  10 &   2045 \\
Validation &   24 &  118 &  316 &   51 &   3 &   512 \\

\bottomrule

\end{tabular}
\caption{Distribution of sentence target levels in training and validation sets.}
\label{tab:unsupervised_dataset_cefr_sample_counts_combined}
\end{table}
\subsection{Supervised Baseline Training}

For the supervised baseline model training, the batch size was 16, and the optimizer used was Adam \citep{kingma2017adam} with a learning rate of $3e-5$, weight decay of 0.1 and 383 warm up steps. The fine-tuning continued for 10 epochs, and a checkpoint with the highest SARI \citep{xu-etal-2016-optimizing} score on the validation set was chosen as the final model.
\subsection{Inference Settings}

We perform inference for open-weight LLMs and the supervised baseline model on NVIDIA GeForce RTX 4090 GPU, with 24 GB of memory, through the Transformers library \citep{wolf-etal-2020-transformers}. For closed-weight models, we use OpenAI APIs for inference. In terms of the generation hyper-parameters, we specify the model to generate maximum output length of 100 tokens, and implement Nucleus Sampling \citep{Holtzman2020The} with a probability of 0.9, a temperature of 1.0. The remaining inference hyper-parameters are set to the default values in Hugging Face and OpenAI APIs. 

To ensure robustness and account for variability in generation, we perform each inference run with three different random seeds. After that, we calculate the average and standard deviation results for all metrics.

\section{CEFR Level Assessment Performance}
\label{app:CEFR-Level-Assessment-Performance}

Table \ref{tab:GPT4_comprission} provides a comparative analysis of CEFR level assessment performance on the publicly available CEFR-SP test set from SCoRE and Wiki-Auto. This comparison compares various LLMs including the GPT-4 model, OpenChat-3.5, and Llama-3-8B-Instruct, as well as the README sentence-level CEFR level estimator \citep{naous2023readme}, and the CEFR sentence-based estimator used in this study \citet{arase-etal-2022-cefr}. The results indicate that the performance of these models was lower than that achieved by the model developed by \citet{arase-etal-2022-cefr}. Moreover, Tables \ref{tab:GPT4_prompt}, \ref{tab:Llama-3-8B-Instruct_prompt} and \ref{tab:OpenChat-3.5_prompt} show the prompt provided to the GPT-4, Llama-3-8B-Instruct and OpenChat-3.5 models for the assessment, respectively.
\begin{table}[h]
\centering
\footnotesize

\setlength{\tabcolsep}{1.3pt}
\begin{tabular}{@{}lccccccc}    
\toprule
Model& A1 & A2 & B1 & B2 & C1 & C2 & Average\\
\toprule

GPT4 &  0.60&  0.73&  0.63&  0.77&  0.67&  0.18&  0.60\\

OpenChat-3.5 &  0.46 &  0.69&  0.78&  0.53&  0.07&  0.00&  0.42 \\

Llama-3-8B-Instruct &  0.05 &  0.07&  0.76&  0.58&  0.00&  0.00&  0.24 \\

README  &  0.63&   0.69 &  0.59& 0.53&  0.36&  0.50&  0.55\\
\midrule
\citet{arase-etal-2022-cefr}  & \textbf{0.82} & \textbf{0.83} & \textbf{0.87} & \textbf{0.87} & \textbf{0.89} & \textbf{0.93} & \textbf{0.87}\\
\bottomrule
\end{tabular}
\caption{Comparison of Macro-F1 scores (\%) across CEFR levels for various LLMs, the README estimator, and the CEFR sentence-based estimator used in this study, evaluated on the publicly available CEFR-SP test set from SCoRE and Wiki-Auto.}
\label{tab:GPT4_comprission}
\end{table}

\begin{table*}[b]
\centering \small
\begin{tabular}{>{\raggedright\arraybackslash}p{0.2\linewidth}p{0.7\linewidth}}
\toprule
\textbf{Role} & \textbf{Content} \\
\midrule
System & Assess the CEFR level (A1, A2, B1, B2, C1, C2) required for an English learner to read and comprehend the provided sentence. Then, return the CEFR level in JSON format \{'cefr\_level': '<level>'\}.\\
& \textbf{A1}: Can understand very short, simple texts a single phrase at a time, picking up familiar names, words, and basic phrases and rereading as required. \\
& \textbf{A2}: Can understand short, simple texts containing the highest frequency vocabulary, including a proportion of shared international vocabulary items. \\
& \textbf{B1}: Can read straightforward factual texts on subjects related to their field of interest with a satisfactory level of comprehension. \\
& \textbf{B2}: Can read with a large degree of independence, adapting style and speed of reading to different texts and purposes, and using appropriate reference sources selectively. Has a broad active reading vocabulary, but may experience some difficulty with low-frequency idioms. \\
& \textbf{C1}: Can understand in detail lengthy, complex texts, whether or not these relate to their own area of speciality, provided they can reread difficult sections. \\
& \textbf{C2}: Can understand virtually all types of texts including abstract, structurally complex, or highly colloquial literary and non-literary writings. \\
\midrule
User & Sentence: [Sentence] \\
\bottomrule
\end{tabular}
\caption{Prompt provided to the GPT-4 model to assess readability assessment performance.}
\label{tab:GPT4_prompt}
\end{table*}

\begin{table*}[t]
\centering \small
\begin{tabular}{>{\raggedright\arraybackslash}p{0.2\linewidth}p{0.7\linewidth}}
\toprule
\textbf{Role} & \textbf{Content} \\
\midrule
System & Assess the CEFR level (A1, A2, B1, B2, C1, C2) required for an English learner to read and comprehend the provided sentence. Then, ONLY return the determined CEFR level.\\
& \textbf{A1}: Can understand very short, simple texts a single phrase at a time, picking up familiar names, words, and basic phrases and rereading as required. \\
& \textbf{A2}: Can understand short, simple texts containing the highest frequency vocabulary, including a proportion of shared international vocabulary items. \\
& \textbf{B1}: Can read straightforward factual texts on subjects related to their field of interest with a satisfactory level of comprehension. \\
& \textbf{B2}: Can read with a large degree of independence, adapting style and speed of reading to different texts and purposes, and using appropriate reference sources selectively. Has a broad active reading vocabulary, but may experience some difficulty with low-frequency idioms. \\
& \textbf{C1}: Can understand in detail lengthy, complex texts, whether or not these relate to their own area of speciality, provided they can reread difficult sections. \\
& \textbf{C2}: Can understand virtually all types of texts including abstract, structurally complex, or highly colloquial literary and non-literary writings. \\
\midrule
User & Sentence: [Sentence]. CEFR level:  \\
\bottomrule
\end{tabular}
\caption{Prompt provided to the Llama-3-8B-Instruct model to assess readability assessment performance.}
\label{tab:Llama-3-8B-Instruct_prompt}
\end{table*}

\begin{table*}[t]
\centering \small
\begin{tabular}{>{\raggedright\arraybackslash}p{0.2\linewidth}p{0.7\linewidth}}
\toprule
\textbf{Role} & \textbf{Content} \\
\midrule
User & Assess the CEFR level (A1, A2, B1, B2, C1, C2) required for an English learner to read and comprehend the provided sentence. Then, return the CEFR level.\\
& \textbf{A1}: Can understand very short, simple texts a single phrase at a time, picking up familiar names, words, and basic phrases and rereading as required. \\
& \textbf{A2}: Can understand short, simple texts containing the highest frequency vocabulary, including a proportion of shared international vocabulary items. \\
& \textbf{B1}: Can read straightforward factual texts on subjects related to their field of interest with a satisfactory level of comprehension. \\
& \textbf{B2}: Can read with a large degree of independence, adapting style and speed of reading to different texts and purposes, and using appropriate reference sources selectively. Has a broad active reading vocabulary, but may experience some difficulty with low-frequency idioms. \\
& \textbf{C1}: Can understand in detail lengthy, complex texts, whether or not these relate to their own area of speciality, provided they can reread difficult sections. \\
& \textbf{C2}: Can understand virtually all types of texts including abstract, structurally complex, or highly colloquial literary and non-literary writings. \\
\midrule
User & Sentence: [Sentence] \\
\bottomrule
\end{tabular}
\caption{Prompt provided to the OpenChat-3.5 model to assess readability assessment performance.}
\label{tab:OpenChat-3.5_prompt}
\end{table*}

\section{Fluency Evaluation}
\label{app:Fluency-Evaluation}

Following \citep{krishna-etal-2020-reformulating,MUCOCO2021}, we evaluate the fluency of generated text using a RoBERTa-large model that has been fine-tuned on the CoLA corpus \citep{warstadt2018neural}. This model evaluates each generated output for grammatical correctness. The final fluency score is then determined by averaging these evaluations across all outputs, providing an overall measure of the text's grammatical quality. Table~\ref{tab:result-by-prompts-with-fluency} shows that all models achieve high fluency scores, independently of the prompt used. Most of them (i.e.\ GPT-4, GPT-3.5-turbo-instruct, Llama-3-8B-Instruct, and Openchat) achieve near-perfect fluency scores. Flan-T5 models also obtain high fluency, with scores between 0.89 and 0.99.

\begin{table*}[p]
\centering
\small
\begin{tabular}{@{}lccc|ccc@{}}
\toprule

 Models &\textbf{$\rho$ (↑)}& \textbf{Accuracy (\%) (↑)}&\textbf{RMSE (↓)} &\textbf{Fluency (↑)}  &\textbf{STS (\%) (↑)}& \textbf{Copies (\%) (↓)}\\
 \midrule
\rowcolor{Gray}\multicolumn{7}{c}{\textit{Baselines}} \\
\midrule
\textbf{COPY} & 0 & 24.15 & 2.45 & 0.99 &  100 & 100 \\
\textbf{SUPERVISED} & 0.01±02 & 39.43±1.44 & 2.08±0.04 & 0.98±0.01 & 84.51±0.75 &12.69±0.91 \\
 \midrule
\rowcolor{Gray}\multicolumn{7}{c}{\textit{P1: target level}} \\
\midrule
Flan-T5-small  &-0.01±0.02&34.69±0.76 & 2.21±0.01 & 0.91±0.02 & 86.70±0.58 & 12.97±1.12 \\
Flan-T5-base  &-0.01±0.01&28.01±0.32 & 2.38±0.02 & 0.97±0.01 & 95.41±0.41 & 30.15±2.24 \\
Flan-T5-large  &-0.00±0.01&28.04±0.49 & 2.36±0.01 & 0.98±0.01 & 96.40±0.25 & 25.68±1.58 \\
Flan-T5-xl  &-0.00±0.01&27.96±0.89 & 2.38±0.01 & 0.99±0.01 & 97.17±0.20 & 30.09±1.39 \\
GPT-3.5-Turbo-Instruct  &0.05±0.01** &39.87±1.03 & 2.01±0.03 & 0.99±0.00 & 85.12±0.36 & 0.00±0.00 \\
GPT-4-Turbo  &  \underline{0.11±0.01**}&\underline{51.68±0.79} & \underline{1.67±0.02} & 0.99±0.01 & 76.96±0.33 & 0.00±0.00 \\
OpenChat\_3.5  &-0.01±0.02&42.87±0.29 & 1.96±0.01 & 0.98±0.01 & 85.35±0.28 & 0.11±0.14 \\

Llama-3-8B-Instruct &0.07 ± 0.00** &44.81±1.57 & 1.89±0.03 & 0.99±0.01 & 84.32±0.23 & 0.58±0.37 \\

Mistral-7B-Instruct-v0.2 &0.06±0.01** &41.40±0.83 & 1.98±0.01 & 0.98±0.01 & 83.63±0.20 & 0.00±0.00 \\

\midrule

\rowcolor{Gray}\multicolumn{7}{c}{\textit{P2: target level + description}} \\

\midrule
Flan-T5-small  & 0.02±0.03 &33.25±0.92 & 2.23±0.03 & 0.91±0.01 & 86.99±0.40 & 13.13±1.29 \\
Flan-T5-base  &-0.01±0.01&27.60±0.70 & 2.39±0.02 & 0.97±0.01 & 95.54±0.45 & 29.54±2.27 \\
Flan-T5-large  &0.01±0.02&28.29±0.88 & 2.36±0.01 & 0.98±0.01 & 96.19±0.35 & 25.96±3.25 \\
Flan-T5-xl  &0.01±0.01&28.49±1.03 & 2.37±0.02 & 0.99±0.01 & 96.95±0.33 & 28.43±1.05 \\
GPT-3.5-Turbo-Instruct &0.08±0.02**&41.51±0.80 & 1.98±0.02 & 0.99±0.01 & 84.95±0.23 & 0.00±0.00 \\
GPT-4-Turbo &\underline{0.18±0.02**}&\underline{52.04±0.69} & \textbf{1.66±0.01} & 0.99±0.01 & 77.24±0.25 & 0.00±0.00 \\
OpenChat\_3.5 & 0.03±0.02&45.08±1.08 & 1.90±0.02 & 0.98±0.01 & 83.04±0.41 & 0.17±0.11 \\

Llama-3-8B-Instruct &0.17 ± 0.01** &46.33±1.01 & 1.83±0.01 & 0.99±0.01 & 83.29±0.35 &0.33±0.19 \\

Mistral-7B-Instruct-v0.2  &0.09±0.01**&41.65±0.88 & 1.96±0.01 & 0.98±0.01 & 83.45±0.31 & 0.00±0.00 \\

\midrule

\rowcolor{Gray}\multicolumn{7}{c}{\textit{P3: target level + examples}} \\

\midrule
Flan-T5-small  &-0.00 ± 0.01&38.79±0.89 & 2.12±0.02 & 0.89±0.01 & 78.54±0.49 & 6.48±1.28 \\
Flan-T5-base  &-0.01 ± 0.01 &29.01±0.90 & 2.35±0.01 & 0.95±0.01 & 93.62±0.32 & 22.94±1.42 \\
Flan-T5-large  &-0.01 ± 0.00&28.43±0.40 & 2.36±0.01 & 0.98±0.01 & 95.75±0.17 & 23.30±3.17 \\
Flan-T5-xl  &-0.01 ± 0.01&28.10±0.55 & 2.37±0.01 & 0.99±0.00 & 96.86±0.26 & 31.78±1.35 \\
GPT-3.5-Turbo-Instruct & 0.08 ± 0.00**&42.12±0.67 & 1.98±0.01 & 0.99±0.01 & 84.97±0.55 & 0.00±0.00 \\
GPT-4-Turbo  &\underline{0.15 ± 0.00**}&\underline{50.73±1.17} & \underline{1.69±0.01} & 0.99±0.00 & 78.13±0.19 & 0.00±0.00 \\
OpenChat\_3.5  & 0.06 ± 0.01*&43.11±0.68 & 1.94±0.01 & 0.98±0.01 & 84.63±0.27 & 0.42±0.27 \\

Llama-3-8B-Instruct &0.11 ± 0.02** &46.35±0.73 & 1.84±0.01 & 0.99±0.00 & 83.46±0.38 &0.61±0.31 \\

Mistral-7B-Instruct-v0.2  &0.08 ± 0.01**&40.23±0.64 & 2.03±0.01 & 0.98±0.00 & 85.66±0.28 & 0.00±0.00 \\

\midrule
\rowcolor{Gray}\multicolumn{7}{c}{\textit{P4: target level + description + examples}} \\

\midrule
 
Flan-T5-small  & -0.02 ± 0.00 & 38.51±1.02 & 2.13±0.01 & 0.89±0.01 & 77.83±0.70 & 7.70±1.08 \\
Flan-T5-base  &-0.01 ± 0.02&30.34±1.31 & 2.33±0.03 & 0.96±0.01 & 91.63±0.28 & 19.59±1.21 \\
Flan-T5-large  &-0.01 ± 0.01 &29.01±0.53 & 2.35±0.01 & 0.98±0.01 & 95.51±0.27 & 22.58±1.71 \\
Flan-T5-xl  &-0.01 ± 0.00&28.70±0.69 & 2.36±0.02 & 0.98±0.01 & 96.57±0.24 & 28.90±1.84 \\
GPT-3.5-Turbo-Instruct  &0.08 ± 0.01**&43.34±1.15 & 1.93±0.02 & 0.99±0.01 & 84.40±0.23 & 0.00±0.00 \\
GPT-4-Turbo  & 0.18 ± 0.01**&\textbf{52.45±0.96} & \textbf{1.66±0.01} & 0.99±0.01 & 77.81±0.27 & 0.00±0.00 \\
OpenChat\_3.5  &0.06 ± 0.01*&44.00±0.80 & 1.91±0.02 & 0.99±0.01 & 83.74±0.38 & 0.25±0.18 \\

Llama-3-8B-Instruct &\textbf{0.19 ± 0.01**} & 48.21±1.23 & 1.80±0.02 &0.99±0.01 &82.86±0.16 &0.50±0.19 \\

Mistral-7B-Instruct-v0.2  & 0.09 ± 0.00**&42.01±0.76 & 1.98±0.01 & 0.98±0.01 & 84.20±0.21 & 0.00±0.00 \\

\bottomrule

\end{tabular}
\caption{Results on the test set, across the four prompts. Metrics: Spearman's $\rho$ , Accuracy, RMSE, Fluency, STS, and Copy Percentage. Boldface indicates the best overall performance across all prompts, while underlined results represent the best performance for each individual prompt. Significance: * (p < 0.01), ** (p < 0.001).}
\label{tab:result-by-prompts-with-fluency}
\end{table*}

\section{Meaning Preservation Supplemental Results}
 
\label{app:bertscore-alignscore-results}

Table~\ref{tab:result-by-prompts-with-bertscore-alignscore} presents the BERTScore and AlignScore results obtained by all selected models in our test set, with the different designed prompts. 

\paragraph{BERTScore \citep{bert-score}:} We compute the BERTScore precision and F1 scores for the predictions against the source sentences. The BERTScore is calculated using Huggingface's implementation with lang="en" and rescale\_with\_baseline=True.

\paragraph{AlignScore \citep{zha-etal-2023-alignscore}:} We compute the AlignScore for the predictions against the source sentences. The score is calculated using the original implementation \url{https://github.com/yuh-zha/AlignScore} with the following settings: model='roberta-base' and evaluation\_mode='nli'.



\begin{table*}[p]
\centering
\small
\begin{tabular}{@{}lccc@{}}
\toprule

 Models &\textbf{BERTScore \textit{precision} (\%)(↑)}& \textbf{BERTScore \textit{f1} (\%)(↑)}&\textbf{AlignScore (\%)(↑)} \\
 \midrule
\rowcolor{Gray}\multicolumn{4}{c}{\textit{Baselines}} \\
\midrule
\textbf{COPY} & 100& 100 & 99.88  \\
\textbf{SUPERVISED} & 74.99±0.89 & 65.15±0.98 & 80.56±1.09 \\
 \midrule
\rowcolor{Gray}\multicolumn{4}{c}{\textit{P1: target level}} \\
\midrule
Flan-T5-small  &73.76±0.83& 66.13±0.75 & 73.57±1.69  \\
Flan-T5-base  &87.50±0.77& 84.25±0.84 & 87.09±1.44  \\
Flan-T5-large  &88.56±0.42& 84.91±0.57 & 93.10±0.42 \\
Flan-T5-xl  &90.29±0.29& 87.29±0.44 & 94.06±0.31  \\
GPT-3.5-Turbo-Instruct  &62.78±0.55&58.38±0.48 & 83.62±0.68 \\
GPT-4-Turbo  &  49.23±0.40& 46.99±0.43 & 71.87±0.60 \\
OpenChat\_3.5  &65.05±0.23& 62.04±0.23 & 83.00±0.55  \\

Llama-3-8B-Instruct &65.65±0.47& 57.93±0.43 & 80.46±0.74  \\

Mistral-7B-Instruct-v0.2 &55.73±0.41& 53.03±0.39 & 79.91±0.39 \\

\midrule

\rowcolor{Gray}\multicolumn{4}{c}{\textit{P2: target level + description}} \\

\midrule
Flan-T5-small  &  73.62±0.57& 66.41±0.68 & 72.98±1.45 \\
Flan-T5-base  & 87.54±0.91& 84.49±1.10 & 86.77±1.88 \\
Flan-T5-large  & 88.48±0.63& 84.42±0.87 & 92.92±0.64 \\
Flan-T5-xl  & 89.90±0.44& 86.62±0.62 & 94.00±0.22 \\
GPT-3.5-Turbo-Instruct & 62.68±0.33& 58.20±0.39 & 83.84±0.66 \\
GPT-4-Turbo & 49.76±0.48& 47.01±0.37 & 71.98±0.84 \\
OpenChat\_3.5 &  61.55±0.53& 58.01±0.52 & 81.56±0.37 \\

Llama-3-8B-Instruct  &64.11±0.52& 56.32±0.50 & 80.46±0.36 \\

Mistral-7B-Instruct-v0.2  & 56.29±0.95& 52.97±0.66 & 79.55±0.84 \\

\midrule

\rowcolor{Gray}\multicolumn{4}{c}{\textit{P3: target level + examples}} \\

\midrule
Flan-T5-small   &62.99±0.45& 54.53±0.39 & 59.03±1.53\\
Flan-T5-base   &84.72±0.40& 80.85±0.48 & 82.53±0.93 \\
Flan-T5-large  &87.59±0.30& 83.46±0.32 & 92.51±0.58 \\
Flan-T5-xl   &90.25±0.30& 86.90±0.28 & 94.14±0.21 \\
GPT-3.5-Turbo-Instruct  &63.32±0.52& 58.33±0.52 & 83.56±0.58 \\
GPT-4-Turbo   &51.37±0.54& 48.44±0.40 & 73.77±0.75 \\
OpenChat\_3.5   &64.14±0.43& 60.64±0.41 & 82.32±0.51 \\

Llama-3-8B-Instruct  &63.87±0.46& 55.95±0.46 & 80.37±0.82 \\

Mistral-7B-Instruct-v0.2   &59.61±0.51& 56.36±0.55 & 82.29±0.66 \\

\midrule
\rowcolor{Gray}\multicolumn{4}{c}{\textit{P4: target level + description + examples}} \\

\midrule
 
Flan-T5-small  &61.84±1.14& 53.75±1.15 & 58.23±2.36 \\
Flan-T5-base   &81.41±0.63& 76.86±0.61 & 78.20±0.78 \\
Flan-T5-large  &87.08±0.58& 82.73±0.67 & 92.09±0.92 \\
Flan-T5-xl  &89.51±0.45& 85.82±0.45 & 93.85±0.50 \\
GPT-3.5-Turbo-Instruct   &63.13±0.33& 57.32±0.25 & 83.76±0.62 \\
GPT-4-Turbo   &51.06±0.21& 47.59±0.25 & 73.03±0.59 \\
OpenChat\_3.5  &63.33±0.72& 59.17±0.61 & 82.05±0.31 \\

Llama-3-8B-Instruct  &63.80±0.60& 55.28±0.47 & 80.25±1.04\\

Mistral-7B-Instruct-v0.2   &58.66±0.25& 54.59±0.28 & 81.29±0.38 \\

\bottomrule

\end{tabular}
\caption{BERTScore and AlignScore results on the test set, across the four prompts. }
\label{tab:result-by-prompts-with-bertscore-alignscore}
\end{table*}

\section{Human Evaluation Details}
\subsection{Readability Assessment}
\label{app:Human-Readability-Assessment-Details}
\begin{figure*}[tb]
    \centering
  
    \begin{subfigure}[b]{0.48\linewidth}
        \centering
        \includegraphics[width=\linewidth]{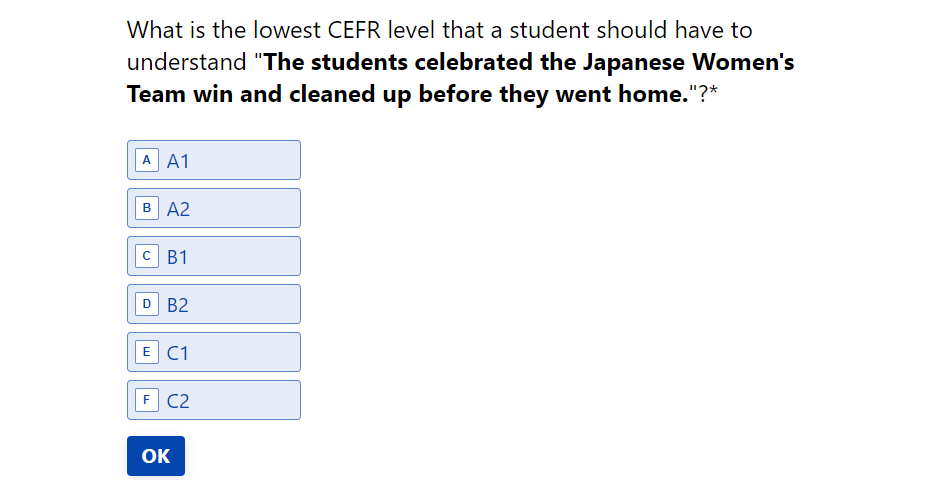}

        \label{fig:RAScreenshot}
    \end{subfigure}
    \hfill
    \begin{subfigure}[b]{0.48\linewidth}
        \centering
        \includegraphics[width=\linewidth]{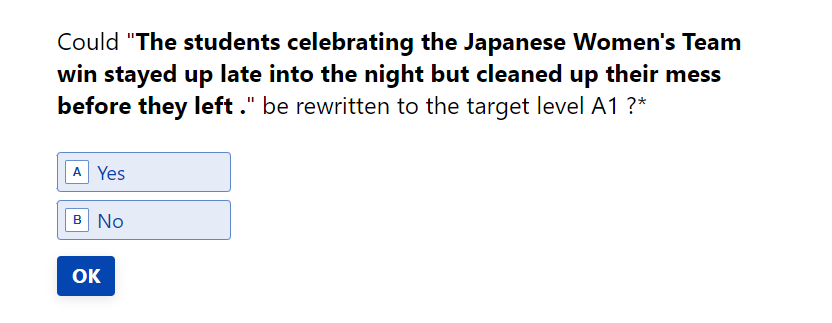}

        \label{fig:RAScreenshot2}
    \end{subfigure}
   
    \caption{Screenshot of the Human Readability Assessment}
    \label{fig:RAScreenshot3}
\end{figure*}

The human readability assessment was conducted by three evaluators with ESL experience, each having varying levels of experience: one native speaker with 1 year of experience, one non-native speaker with 3 years of experience, and one native speaker with 25 years of experience. To ensure that only qualified evaluators who are  familiar with, and understand the CEFR descriptor scheme are included, we implemented a qualification task. This task comprised 30 sentences annotated with CEFR levels from the CEFR-SP dataset~\citep{arase-etal-2022-cefr}. Evaluators had to achieve at least 80\% accuracy on this set to qualify. We measured inter-annotator agreement using two metrics: Percentage of Majority Agreement (i.e., the proportion of instances that had a clear majority vote) and Krippendorff's alpha (ordinal level). The evaluators achieved 96.66\% and 88.88\% Majority Agreement in the qualification and rating tasks, respectively. Krippendorff's alpha indicated a high level of agreement ($\alpha = 0.85$) in the qualification task and moderate agreement ($\alpha = 0.43$) in the rating task. Additionally, each evaluator’s agreement with the annotated examples in the qualification task was measured using Krippendorff’s alpha ($\alpha = 0.86$, $\alpha = 0.86$, $\alpha = 0.87$), indicating strong agreement. Figure~\ref{fig:RAScreenshot3} shows screenshots of the human readability assessment.

\subsection{Meaning Evaluation}
\label{app:Human-Meaning-Evaluation-Details}
\begin{figure*}[tb]
    \centering
    \includegraphics[width=\linewidth]{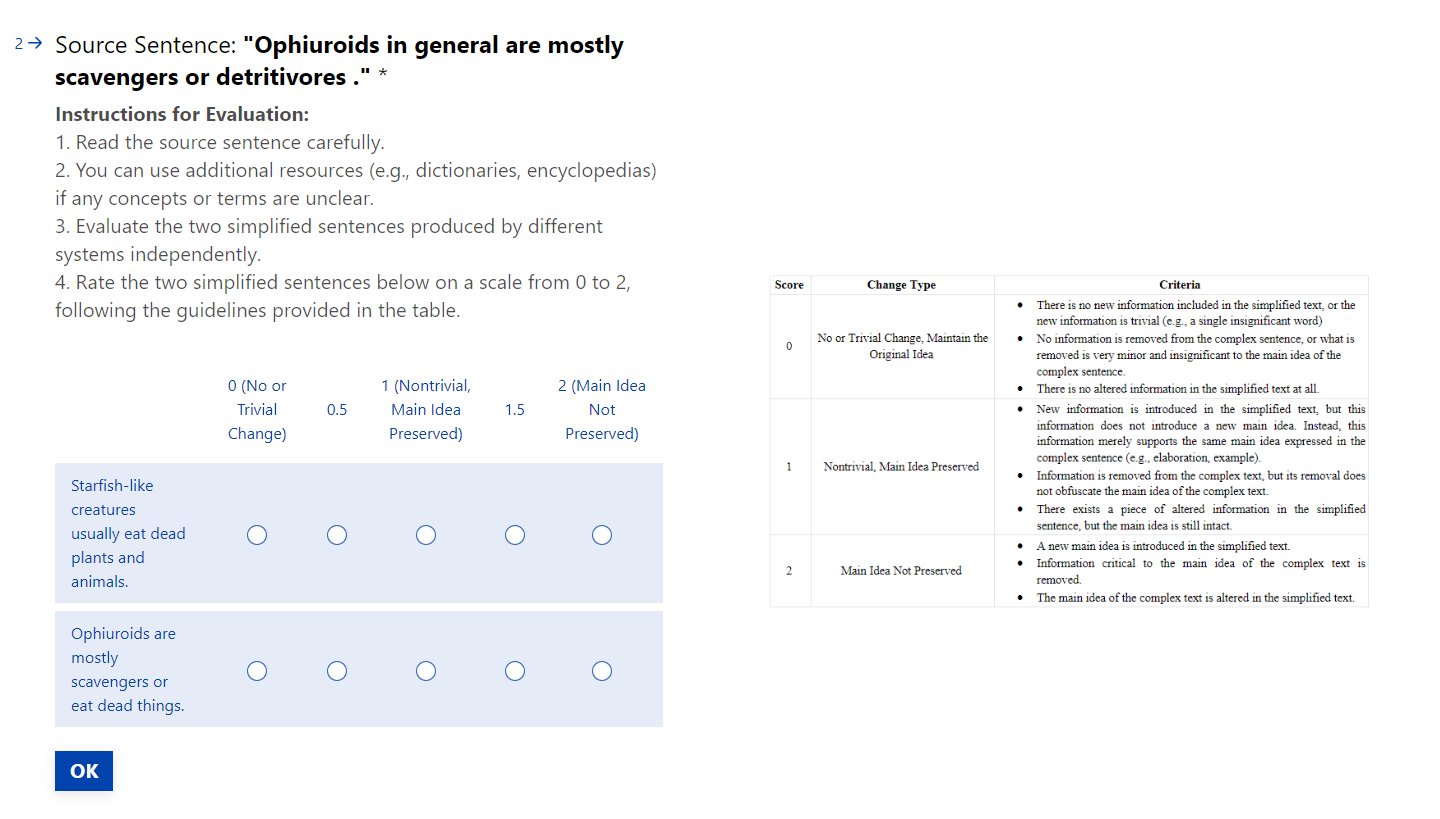}
    \caption{Screenshot of the Human Meaning Evaluation}
    \label{fig:HMPScreenshot}
\end{figure*}

\begin{figure*}[h]
    \centering
    \includegraphics[width=\linewidth]{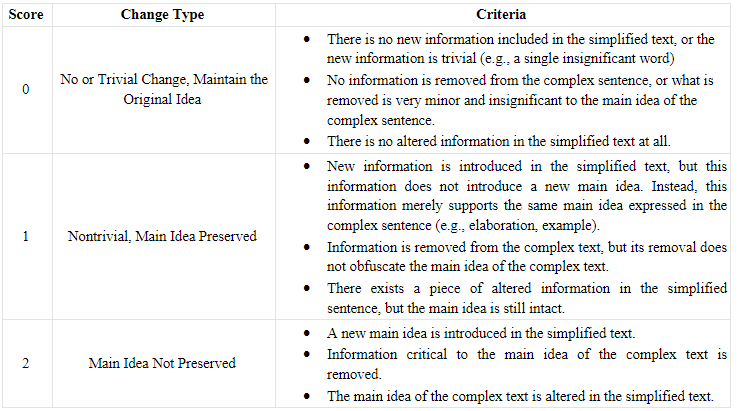}
    \caption{Criteria Provided to the Evaluators for the Meaning Evaluation }
    \label{fig:HMPcriteria}
\end{figure*}

The human meaning evaluation was conducted by three fluent English speakers: two native speakers and one non-native but fluent speaker. To ensure that only evaluators who are qualified and understand the scoring scheme are included, we implemented a qualification task. This task consisted of 20 sentence pairs with substantial agreement ($\alpha = 0.79$) among three of the authors. Evaluators were required to achieve at least ($\alpha = 0.75$) average agreement with the authors on this set. The selected participants achieved Krippendorff's alpha scores of .79, .82, and 82, based on an ordinal level of measurement. We also determined inter-annotator agreement using Percentage of Majority Agreement and Krippendorff's alpha (ordinal level). The evaluators achieved 90.0\% and 82.2\% Majority Agreement in the qualification and rating tasks, respectively. Krippendorff's alpha indicated a high level of agreement ($\alpha = 0.86$) in the qualification task and moderate agreement ($\alpha = 0.31$) in the rating task. Figure~\ref{fig:HMPScreenshot} and Figure~\ref{fig:HMPcriteria} display screenshots of the human meaning evaluation and the evaluation criteria provided to the evaluators.

\end{document}